\documentclass[11pt]{article}
\RequirePackage{silence}
\usepackage{amssymb}
\usepackage{longtable}
\usepackage{booktabs}
\usepackage{booktabs}
\usepackage{makecell}
\usepackage{array}
\usepackage{ragged2e}
\usepackage{tabularray}
\UseTblrLibrary{booktabs}
\usepackage{fontspec}
\newfontfamily\banglafont{kalpurush.ttf}[
    Path=./font/, 
    Script=Bengali,
    SizeFeatures={Size=11}
]
\newcommand{\bangla}[1]{{\banglafont #1}}

\usepackage{listings}
\usepackage{xcolor}

\definecolor{pykeyword}{HTML}{0000FF}
\definecolor{pystring}{HTML}{A31515}
\definecolor{pycomment}{HTML}{008000}
\definecolor{pydecorator}{HTML}{2B91AF}
\definecolor{transgray}{HTML}{6C757D} % Muted gray for translations

\usepackage[preprint]{acl}
\usepackage{times}
\usepackage{latexsym}
\usepackage{blindtext}
\usepackage{amsmath}
\usepackage[T1]{fontenc}
\usepackage{tcolorbox}
\usepackage{xcolor}
\usepackage{enumitem}
\usepackage{amssymb}
\usepackage{booktabs}
\usepackage{multirow}
\usepackage{soul}
\usepackage[utf8]{inputenc}

\usepackage{microtype}

\usepackage{inconsolata}

\usepackage{graphicx}

\title{Beyond Clean Text: Evaluating Encoder and Decoder Robustness for Bangla Event Detection in Noisy Text}

\author{
    Tanvir Ahmed Sijan\textsuperscript{1, \dag} \quad
    S. M Golam Rifat\textsuperscript{2} \quad
    \textbf{Nayeemul Islam\textsuperscript{3}} \quad
 \\
    \textbf{Md. Musfique Anwar\textsuperscript{1}} \quad
\\
 \textsuperscript{1}Jahangirnagar University, Dhaka, Bangladesh,
 \\
  \textsuperscript{2}Rajshahi University of Engineering \& Technology, Rajshahi, Bangladesh
  \\
\textsuperscript{3}Bangladesh University of Engineering and Technology, Dhaka, Bangladesh,
 \\
\texttt{\{sijantanv, golamrifat, nayeemulislam.eee.buet\}@gmail.com} \\
\texttt{manwar@juniv.edu} \quad
\textsuperscript{\dag}Corresponding author
 }

\usepackage{lipsum}
\begin{document}
\maketitle
\begin{abstract}
Event detection (ED) systems are typically evaluated on clean, curated text, leaving their robustness to real-world noise largely unexplored, particularly for low-resource languages such as Bangla. We introduce a generalized Bangla news event ontology and a benchmark comprising 9,979 annotated sentences across 40 event subtypes, spanning clean news text, real-world Automatic Speech Recognition (ASR) transcripts, and orthographically corrupted text. We systematically evaluate fine-tuned encoder-only models (BanglaBERT and XLM-R) alongside instruction-tuned decoder-only large language models (Llama 3 and Gemma 3). Our results reveal a clear architectural trade-off: encoder models achieve higher performance on clean text but degrade substantially under noise, whereas decoder-only LLMs are markedly more robust, particularly when event triggers are corrupted. We further show that embedding annotation guidelines during instruction tuning establishes a higher performance baseline on noisy text but yields inconsistent reductions in performance degradation across noisy conditions. Finally, model scaling consistently improves the robustness of decoder-only LLMs, while combined training on clean and noisy data serves as an effective regularization strategy that disproportionately benefits encoder architectures, significantly narrowing the robustness gap.

\end{abstract}

\section{Introduction}
The task of event detection (ED) identifies and categorizes events in natural language \citep{ahnStagesEventExtraction2006}. It serves as a foundational component in information retrieval. Extracting structured event frames is critical for downstream applications, particularly in emergency monitoring systems where rapid and accurate information retrieval is paramount. Despite its importance, the majority of event detection research focuses almost exclusively on clean and carefully curated text \citep{wangMAVENMassiveGeneral2020, pouranbenveysehMINIONLargeScaleDiverse2022, yaoLEVENLargeScaleChinese2022}. In real-world applications, data is inherently noisy. Even minor typographical errors or transcription faults can cause traditional event detection systems to become confused and miss vital events entirely.

\begin{figure}[t]
  \centering
  \includegraphics[width=\columnwidth]{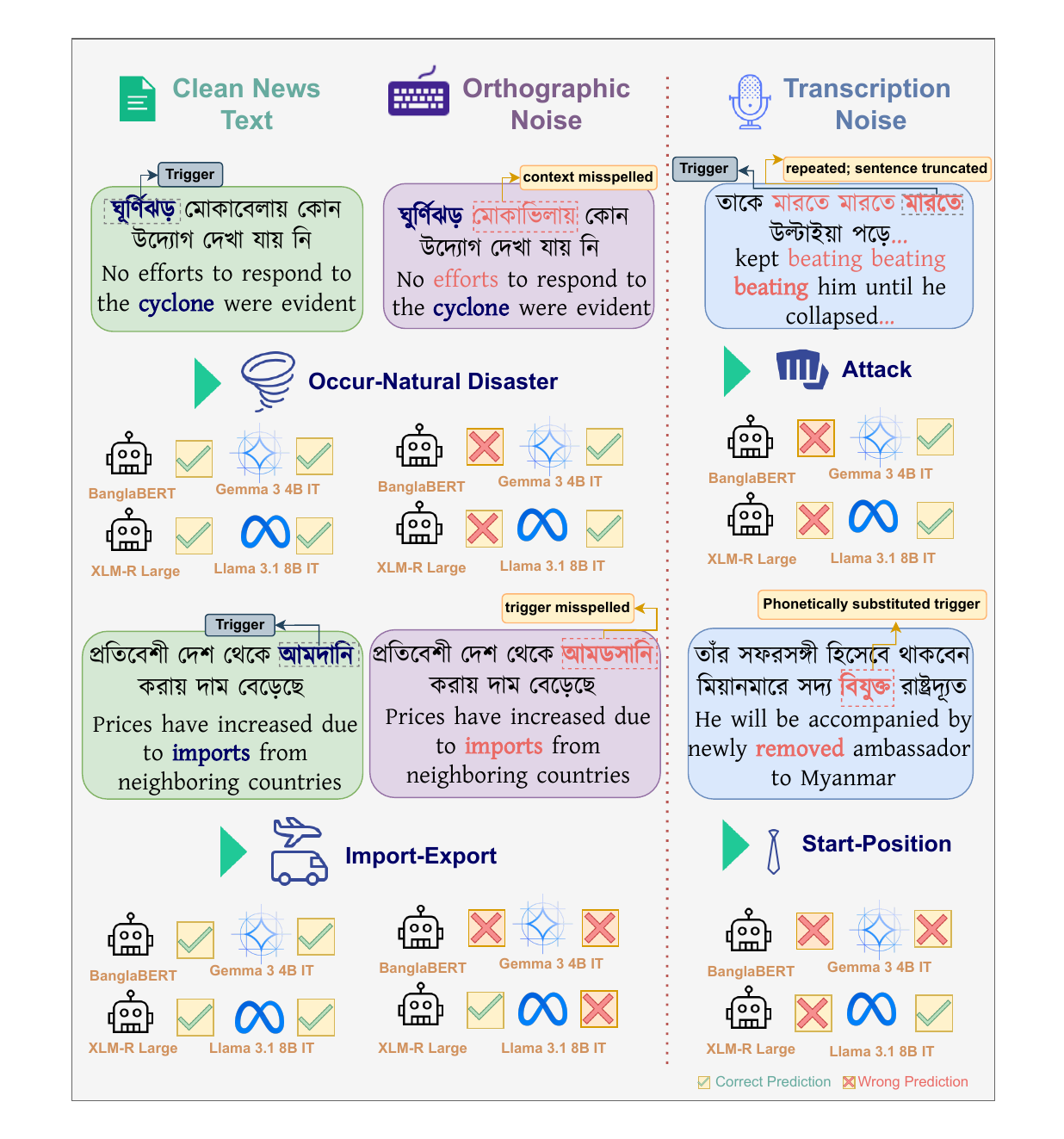}
  \caption{Real-world noise substantially degrades event detection, even when the event trigger itself remains unchanged. Clean examples are drawn from our Clean Test set, while their counterparts are generated using simulated orthographic noise. ASR examples are collected from real Bangla news video transcripts and independently annotated using the same event ontology, introducing naturally occurring transcription errors and out-of-distribution vocabulary and transcription artifacts.}
  \label{fig:prompt_design}
\end{figure}

As a low-resource language, event detection studies in Bangla remain limited. Existing work has primarily focused on clean text and narrow domains, such as violent incidents \citep{khandokarEventDetectionKnowledge2020a, deyUsingMachineLearning2021, alikhandokarTemporalDemographicGeographical2025}, disaster \citep{daveFIRE2020EDNIL2021} or crime-related events \citep{hossainMaskNetEnhancingCrime2025}. Furthermore, there is currently no generalized news event ontology for trigger-based event detection comparable to widely used schemas such as ACE 2005 \citep{walkerchristopherACE2005Multilingual2006a}, thereby limiting the systematic study of the task. 

Even research on noisy Bangla text remains scarce, and the few existing studies have predominantly focused on sentiment analysis \cite{islamSentNoBDatasetAnalysing2021, elahiComparativeAnalysisNoise2024}. Sentiment analysis and event detection pose fundamentally different challenges. While sentiment analysis typically operates at the sequence or document level, event detection requires fine-grained token-level identification and classification of event triggers.

Historically, event extraction has been dominated by BERT-based encoder models \citep{wangMAVENMassiveGeneral2020,pouranbenveysehMINIONLargeScaleDiverse2022, huangTextEEBenchmarkReevaluation2024}. This stands in contrast to the decoder-only models that dominate the current LLM landscape, whose application to structured extraction tasks has traditionally been limited by hallucination and difficulties in generating outputs that conform to predefined structures. However, recent studies have attempted to improve the performance of LLMs on event extraction through techniques such as code-like representations \citep{wangCode4StructCodeGeneration2023}, incorporation of sentence-level contextual information \citep{alMonsurEventDetectionContextAware2026}, and instruction tuning with annotation guidelines to enhance cross-schema generalization \citep{srivastavaInstructionTuningLLMsEvent2025}, while generative formulations have also been explored for complex event argument extraction \citep{sharifExplicitImplicitScattered2024}.

Given these recent methodological advances for adapting LLMs to structured prediction tasks, an important question emerges regarding how these models compare with their BERT-based counterparts under real-world noisy conditions. This question is particularly relevant for Bangla, where the number and scale of available pre-trained encoder models are limited, while recent multilingual LLMs have demonstrated increasingly strong capabilities in the language. 

To address these questions, we annotate 5,320 sentences collected from Bangla newspapers and 4,659 sentences of automatic speech recognition (ASR) transcripts obtained from Bangla news videos within the same domain. We evaluate both encoder-based and decoder-only models for event detection. For decoder-only models, we formulate event detection as a structured code generation task following \citet{wangCode4StructCodeGeneration2023}, with  optionally annotation guidelines embedded in the prompt \citep{sainzGoLLIEAnnotationGuidelines2024, srivastavaInstructionTuningLLMsEvent2025}. In addition, to simulate orthographic noise, we employ the error generation algorithm of \citet{sifatSyntheticErrorDataset2020}, which was developed by analyzing common Bengali writing patterns and typographical behaviors, and evaluate model robustness under varying degrees of noise. In summary, our contributions are as follows:
\begin{itemize}
    \item We develop a generalized news-domain event schema for Bangla event detection and release a dataset comprising 9,979 annotated sentences spanning both clean news text and noisy ASR transcripts, containing 7813 event mentions across 40 event subtypes.
    \item We provide a systematic comparison of encoder-based and decoder-only architectures with varying parameter sizes and degrees of Bangla support under multiple noisy conditions. For decoder-only models, we adopt recent recommendations for structured output generation through instruction tuning and code-based representations.
    \item We present the first comprehensive study of robustness for Bangla event detection under both Real-World ASR-induced and simulated orthographic noise, providing insights into the relative strengths and limitations of modern multilingual LLMs and traditional encoder-based approaches.
\end{itemize}

\begin{figure*}[!ht]
\centering
    \includegraphics[scale=0.62]{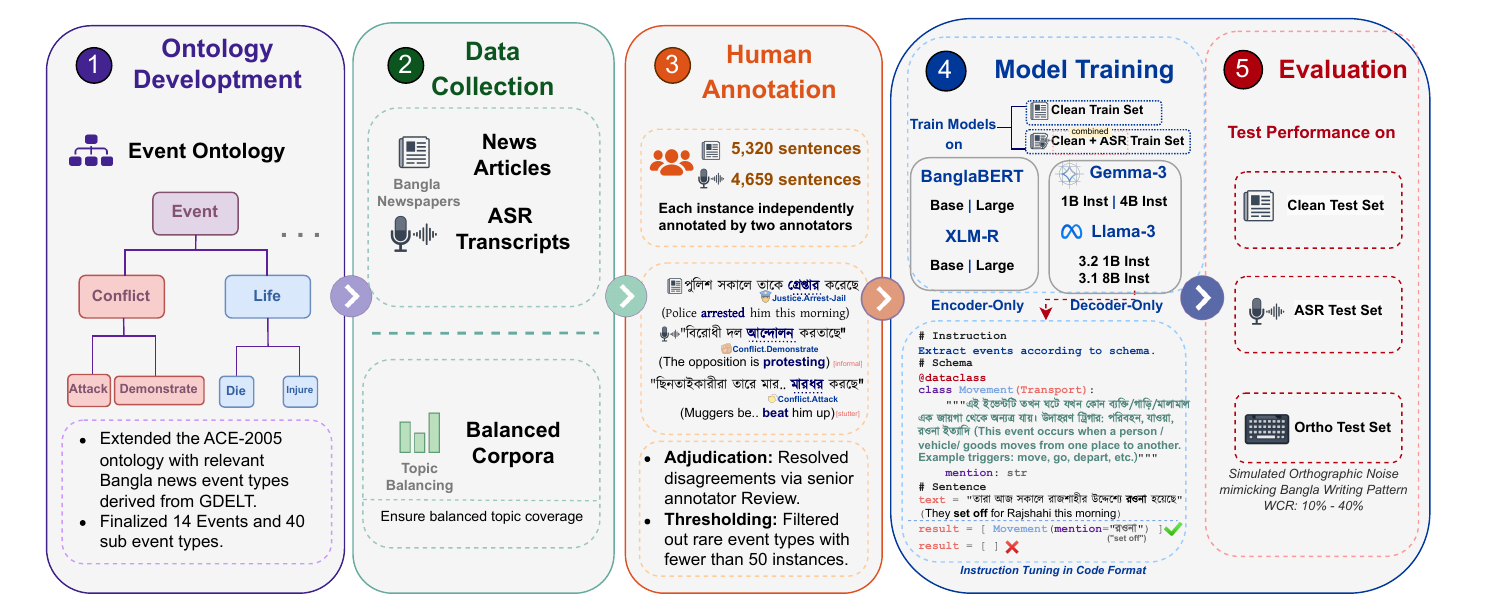}
    \caption{Overview of the dataset construction, training, and evaluation pipeline. We develop a generalized Bangla news event ontology, collect topic-balanced news and ASR corpora, and annotate both using the same event schema. Encoder-only and decoder-only models are trained under two settings: Clean and Combined (Clean + ASR) Training Set. Decoder-only LLMs are instruction-tuned using code-format prompts with and without annotation guidelines embedded as Python docstrings. Robustness is evaluated on the Clean, real-world ASR transcription noise, and simulated orthographic noise test sets.}
    \label{fig:overview}
\end{figure*}

\section{Related Work} 

Early event detection systems relied on feature engineering and statistical learning techniques \citep{ahnStagesEventExtraction2006, patwardhanUnifiedModelPhrasal2009, hongUsingCrossEntityInference2011, liJointEventExtraction2013}. More recently, pretrained Transformer encoders such as BERT have become the dominant paradigm, substantially advancing trigger detection performance \citep{nguyenTrankitLightWeightTransformerbased2021, wangMAVENMassiveGeneral2020, pouranbenveysehMINIONLargeScaleDiverse2022, huangTextEEBenchmarkReevaluation2024}. Alongside these architectural advances, recent work has also focused on developing event detection datasets spanning diverse domains \citep{kimOverviewBioNLP09Shared2009, simsLiteraryEventDetection2019, leFineGrainedEventTrigger2021, yaoLEVENLargeScaleChinese2022} and languages \citep{pouranbenveysehMINIONLargeScaleDiverse2022, touilebEDENDatasetEvent2024}. Large-scale resources such as MAVEN \citep{wangMAVENMassiveGeneral2020}, RAMS \citep{ebnerMultiSentenceArgumentLinking2020}, and the TextEE benchmark \cite{huangTextEEBenchmarkReevaluation2024} have become standard evaluation platforms for modern event detection research.

Compared with the extensive literature on English event detection, Bangla has received considerably less attention. Existing studies have primarily developed task-specific datasets and models targeting individual application domains, including violent incidents \citep{khandokarEventDetectionKnowledge2020a, deyUsingMachineLearning2021, alikhandokarTemporalDemographicGeographical2025}, disasters \citep{daveFIRE2020EDNIL2021}, and crime-related news \citep{hossainMaskNetEnhancingCrime2025}. Moreover, publicly available resources do not provide a generalized news-domain ontology comparable to ACE 2005 \citep{walkerchristopherACE2005Multilingual2006a}, making standardized evaluation across diverse event categories difficult. Studies on noisy Bangla text are likewise limited and have focused primarily on sentence-level tasks such as sentiment analysis \citep{islamSentNoBDatasetAnalysing2021, elahiComparativeAnalysisNoise2024} rather than token-level trigger identification.

The rapid progress of decoder-only LLMs has recently motivated their application to event extraction. Recent studies have explored adapting LLMs to this task through structured code representations \citep{wangCode4StructCodeGeneration2023}, context-aware encoder \citep{alMonsurEventDetectionContextAware2026}, and instruction tuning with annotation guidelines to improve schema understanding and cross-schema generalization \citep{srivastavaInstructionTuningLLMsEvent2025}. Beyond event extraction, annotation guidelines have also been shown to improve other information extraction tasks, including Named Entity Recognition \citep{sainzGoLLIEAnnotationGuidelines2024} and relation extraction \citep{pangGuidelineLearningInContext2023}. However, these methods have almost exclusively been evaluated on clean English benchmarks, leaving the robustness of encoder-only and decoder-only models, as well as the impact of annotation guidelines during instruction tuning, under realistic noisy conditions largely unexplored.

In this work, we address these gaps using Bangla as a representative low-resource language. We introduce a generalized Bangla news event ontology, construct a benchmark comprising clean news articles, real-world ASR transcripts, and simulated orthographic noise, and systematically compare fine-tuned encoder-only and instruction-tuned decoder-only models. We further investigate whether annotation guidelines improve model performance and robustness under noisy conditions.

\section{Benchmark Development}
\subsection{Task Formulation}

We evaluate event detection across two distinct modeling paradigms: Token Classification, for encoder-only architectures and Structured Sequence Generation, for decoder-only large language models. 

Formally, let an input text sequence of length $n$ be defined as $X = [x_1, x_2, \dots, x_n]$. Our ontology consists of a predefined set of event types $\mathcal{E}$. The goal of the event detection system is to extract a set of event instances $Y$ from $X$. Each extracted event instance is defined as a tuple $(i, j, t)$, where $[x_i, \dots, x_j]$ represents the continuous text span that triggers the event, and $t \in \mathcal{E}$ is the predicted event type.  To systematically evaluate the capabilities of different architectures, the extraction of $Y$ is formulated through two paradigms:

\paragraph{Event Detection as Token Classification.} Under this traditional sequence-labeling paradigm, the task is framed as a token-level classification problem. A model processes the input text $X$ and assigns a label $y_k$ to each token. The model must simultaneously perform \textit{Trigger Identification} (locating the boundaries $i$ and $j$ of the trigger span) and \textit{Trigger Classification} (mapping the identified span to an event type $t \in \mathcal{E}$). 

\paragraph{Event Detection as Structured Sequence Generation.} For generative LLMs, the task is reframed as conditional sequence generation. Given the input sentence $X$, a natural language task instruction $I$, and the target event schema $E_t$ for a specific event type $t \in \mathcal{E}$, we construct a unified prompt $P$:
\begin{equation}
    P = [I \oplus E_t \oplus X]
    \label{eq:prompt_formulation}
\end{equation}
where $\oplus$ denotes string concatenation. The instruction $I$ dictates the extraction rules and desired structured output format, while $E_t$ provides the semantic definition of the target event. The model is trained to autoregressively generate the structured representation of $Y$, explicitly handling both the extraction of valid triggers and the rejection of schemas not present in $X$.

\subsection{Ontology Development}
We adopted the widely used ACE 2005 \citep{walkerchristopherACE2005Multilingual2006a} event ontology as the foundation of our schema. However, several event categories frequently appearing in Bangladeshi news, such as natural or manmade disasters, disease outbreaks, festivals, socio-economic events, and certain crime-related events, are not explicitly represented in ACE 2005.

To identify event types relevant to the local news landscape, we analyzed Bangladesh-specific event distributions from GDELT \citep{leetaru2013gdelt} and examined frequently occurring themes in the GKG corpus between 2023 and 2025. Specifically, we extracted and ranked CAMEO event codes and GKG themes associated with Bangladesh to understand the types of events commonly reported in local news.

Based on this analysis, we extended the ACE 2005 ontology with locally relevant event types and initially constructed a schema consisting of 16 top-level event categories and 70 event sub-types. Following the annotation process, event types with fewer than 50 instances were removed to improve class coverage and evaluation reliability. The final ontology comprises 14 main events and 40 sub-events. The ontology details are presented in Appendix \ref{sec:ontology}.

\subsubsection{Data Collection}
\paragraph{News Text} To construct the clean text corpus, we utilized the Bangla News Article Dataset (BNAD) \citep{saadBanglaNewsArticle2024}, which contains category-wise collections of news articles sourced from Bangla Online Newspaper. We collected articles from three major Bangladeshi newspapers: Dhaka Tribune Bangla, The Daily Ittefaq, and Daily Janakantha. Articles were sampled from categories aligned with the event categories in our ontology to ensure balanced coverage across diverse domains. Coverage of topics in the dataset are described in Appendix \ref{sec:ontology}.
\paragraph{ASR Trasncript}
To construct our real-world noisy ASR corpus, we curated publicly available news broadcasts from the official YouTube channels of three of the most popular Bangladeshi television networks: Jamuna TV, Somoy TV, and Independent Television. We extracted the automatically generated Bangla speech transcripts associated with these broadcasts. Utilizing these ASR transcripts provides a naturally occurring source of noisy Bangla text. This data inherently captures colloquial vocabulary, dialectal verb inflections, homophone substitutions, and transcription errors, accurately reflecting the morphological and lexical challenges encountered by event extraction systems in real-world spoken media.

\subsection{Dataset Quality Control}
\paragraph{Annotation Procedure}
To ensure annotation quality and consistency, we first developed a comprehensive annotation guideline based on the proposed ontology. The guideline provided detailed descriptions of fine-grained event types and subtypes, representative trigger example, boundary annotation instructions, and resolutions for common edge cases. Based on these guidelines, we designed a screening test to assess potential annotators. Ultimately, six annotators who successfully passed the screening were recruited. The annotation process was conducted using the academic version of Label Studio \citep{label_studio}.

\paragraph{Inter-Annotator Agreement} Each instance was independently annotated by two annotators. Inter-annotator agreement (IAA), measured using Cohen's $\kappa$, was 0.72, indicating substantial agreement.

\paragraph{Adjudication and Quality Control} Annotation disagreements were resolved through a two-stage adjudication process consisting of peer discussion followed by verification by senior annotators. Annotated Sentences with unresolved ambiguity were excluded from the final dataset. Overall, approximately 18\% of annotated sentences were discarded during the quality control process.

\subsection{Dataset Statistics}
In total, the finalized benchmark corpus comprises 9,979 annotated sentences, including 5,320 sentences from news articles and 4,659 sentences from ASR transcripts, containing a total of 7,813 event mentions. Dataset statistics are summarized in Table \ref{tab:dataset_stats}.
 \begin{table}[htbp]
\centering
\small
\setlength{\tabcolsep}{4pt} % Tightens the space between columns
\begin{tabular}{lccc}
\toprule
\textbf{Source} & \makecell{\textbf{\# Annotated} \\ \textbf{Sentences}} & \makecell{\textbf{\# Event} \\ \textbf{Mentions}} & \makecell{\textbf{Avg. Words} \\ \textbf{/Instance}} \\
\midrule
News Article & 5320 & 4659 & 14.1 \\
ASR Transcript & 4484 & 3407 & 11.5 \\
\midrule
\textbf{Total} & \textbf{9979} & \textbf{7813} & \textbf{12.9} \\
\bottomrule
\end{tabular}
\caption{Dataset Statistics}
\end{table}

\begin{figure}[htbp]
    \centering
    \includegraphics[width=\linewidth]{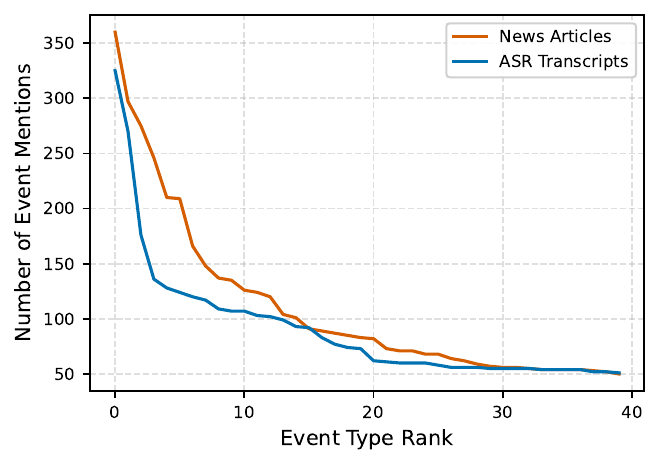}
    \caption{Frequency distribution of the 40 event types in the News and ASR corpora, showing the long-tail nature of both datasets. Rankings are computed independently for each corpus.}
    \label{fig:event_distribution}
\end{figure}

\subsection{Event Class Distribution}
Both the News Text and ASR Transcript subsets exhibit a natural long-tail distribution over the 40 event subtypes, as shown in Figure~\ref{fig:event_distribution}. To maintain evaluation reliability and ensure comparable coverage, event types with fewer than 50 instances were discarded during ontology refinement. As a result, the benchmark preserves the realistic class imbalance inherent in real-world reporting while providing sufficient data support for all retained classes.

\subsection{Orthographic Noise Test Data}
While our ASR corpus captures transcription errors arising from acoustic ambiguity, pronunciation variation, and missing or incorrectly recognized tokens, these differ substantially from the orthographic noise commonly encountered in user-generated Bangla text. Orthographic errors in Bangla are predominantly lexical and character-level phenomena caused by keyboard proximity, phonetic substitutions, spelling mistakes, and the incorrect handling of \textit{Juktakkhar} (consonant conjuncts). Therefore, robustness against ASR errors does not inherently guarantee robustness against typographic noise.

Collecting a large corpus of naturally occurring orthographic errors would require additional annotation effort and label re-verification. Instead, we simulate typographic noise using the alogorithm proposed by \citet{sifatSyntheticErrorDataset2020}, which generates realistic Bangla typing errors based on English QWERTY keyboard patterns, including phonetic substitutions, neighboring key errors, insertion patterns, and \textit{Juktakkhar}-specific rules. This approach also enables controlled evaluation of model robustness under varying noise levels while preserving the original event annotations.

To systematically evaluate model behavior under increasing levels of orthographic degradation, we apply this noise injection process to our Clean Test set, parameterizing the noise injection process using Word Corruption Rate (WCR), defined as $x \in \{10,20,30,40\}$. Noise is injected cumulatively, such that each higher WCR level retains all corruptions introduced at lower levels while introducing additional corrupted words until approximately $x\%$ of the words are modified. The remaining context is left unchanged.

\section{Experiments}
\subsection{Experimental Setup}
\paragraph{Datasets and Evaluation Splits} 
We evaluate model performance and robustness across three distinct test sets: the \textbf{Clean Test} set, comprising standard Bangla news text; the \textbf{ASR Test} set, consisting of automatically generated transcripts from Bangla youtube news videos; and the \textbf{Ortho Test} set, generated from the Clean Test set by simulating realistic orthographic errors at varying Word Corruption Rates (WCRs). Because sentences frequently contain multiple event triggers following a long-tail distribution, standard random splitting risks omitting rare classes from the evaluation sets, which would severely compromise the reliability of our evaluation. To ensure rigorous class-wise representation, we partition the data using an iterative multi-label stratification algorithm \citep{sechidis2011stratification, pmlr-v74-szymański17a}. Using this method, the Clean corpus was partitioned into a 70/15/15 ratio (training/validation/testing), while the ASR corpus was partitioned into a 70/10/20 ratio. The final sentence-level distributions for these datasets are detailed in Table \ref{tab:dataset_stats}.

\begin{table}[ht]
\centering
\begin{tabular}{@{}lrr@{}}
\toprule
\textbf{Dataset} & \textbf{Train \& Val} & \textbf{Test} \\
\midrule
News Text & $4504$ & $816$ \\
ASR Transcripts & $3705$ & $954$ \\
\midrule
Combined (Clean + ASR) & $8209$ & -- \\
Ortho Test  & -- & $816^{\dagger}$ \\
\bottomrule
\end{tabular}
\caption{Raw sentence counts across dataset splits. $^{\dagger}$Indicates the count per discrete WCR level}
\label{tab:dataset_stats}
\end{table}

\paragraph{Models} 
For the encoder-only baselines, we employ BanglaBERT (Base: 110M, Large: 335M) \citep{bhattacharjeeBanglaBERTLanguageModel2022} and XLM-RoBERTa (Base: 270M, Large: 550M) \citep{conneauUnsupervisedCrosslingualRepresentation2020}. BanglaBERT represents the state-of-the-art monolingual pretrained model for Bangla NLP, while XLM-R provides a strong multilingual baseline.

For the generative LLM baselines, we evaluate several open-weight LLMs across different parameter scales and multilingual support. Specifically, we consider models from the Llama family, including Llama 3.1 8B Instruct and Llama 3.2 1B Instruct \citep{dubey2024llama}, together with Google's Gemma 3 4B Instruct and Gemma 3 1B Instruct \citep{gemmateamGemma3Technical2025}.
\paragraph{Model Training}
For  encoder-only architectures, event detection is formulated directly as a token classification task.  For generative LLMs event detection is reformulated as an instruction-following task. We employ a targeted negative sampling (NS) approach. Negative examples provide valuable contrastive supervision for distinguishing event types and suppressing hallucinated predictions \citep{srivastavaInstructionTuningLLMsEvent2025}.
To assess the impact of annotation guidelines on both model performance and robustness to noisy inputs, we consider two independent prompting settings. While both settings provide the model with the input sentence and the target event type, they differ in their schema representation. In the \textbf{w/ Guide} setting, detailed event definitions are embedded directly within the prompt as Python docstrings. In contrast, the \textbf{w/o Guide} setting omits these definitions entirely, requiring the model to perform event detection based solely on the event type name and its underlying parametric knowledge. In both settings, we instruction-tune the LLMs on an augmented training set where each sentence's positive event queries are supplemented with exactly $5$ randomly selected negative samples, yielding a 1:5 positive-to-negative ratio. We use a smaller negative sampling ratio than prior work as a practical trade-off between computational efficiency and exposure to absent event types. Details on prompt format and training setup are described in Appendix \ref{sec:prompt_formats} and \ref{sec:train}. 
\paragraph{Evaluation Protocol}
Following \citet{srivastavaInstructionTuningLLMsEvent2025}, we adopt a one-vs-all inference strategy for generative LLMs, where each test sentence is paired with every event type schema independently, and predictions are aggregated across all 40 types. 

Encoder-only models naturally learn to identify non-event tokens through the \texttt{O} label in BIO tagging, whereas generative LLMs require explicit negative supervision to correctly abstain when the queried event type is absent. The one-vs-all formulation therefore provides a comparable supervision signal across both architectural paradigms, enabling a fair comparison despite their fundamentally different output formulations.

\paragraph{Evaluation Metric}
We select Macro-F1 as our primary evaluation metric. As demonstrated by \citet{alMonsurEventDetectionContextAware2026}, prioritizing Micro-F1 systematically inflates perceived model performance by disproportionately favoring majority classes. Because Macro-F1 computes the metric independently for each class before averaging, it equally weights all categories regardless of their support size.

\begin{table*}[t]
\centering
\resizebox{\textwidth}{!}{%
\begin{tblr}{
  colspec = {|l|l|l|c|c|c|c|c|},
  hlines,
  cell{1}{1} = {r=2}{m},
  cell{1}{2} = {r=2}{m},
  cell{1}{3} = {r=2}{m},
  cell{1}{4} = {c=5}{c},
}
\textbf{Architecture} & \textbf{Model} & \textbf{Training Condition} & \textbf{Macro F1} & & & & \\
 & & & \textbf{Clean Test} & \textbf{ASR Test} & \textbf{Ortho Test (Avg.)} & \textbf{$\Delta_{\text{ASR}}$} & \textbf{$\Delta_{\text{Ortho}}$} \\

\SetCell[r=12]{m} Decoder-Only
& \SetCell[r=3]{m} Gemma 3 4B IT
& Clean w/ Guide
& 60.57 & 59.53 & 53.66 & 1.04 & 6.91 \\

&
&
Clean w/o Guide
& 55.19 & 53.94 & 50.58 & 1.25 & 4.61 \\

&
&
Combined w/ Guide
& 65.60 & 65.43 & 59.14 & 0.17 & 6.46 \\

&
\SetCell[r=3]{m} Gemma 3 1B IT
& Clean w/ Guide
& 52.27 & 48.81 & 45.11 & 3.46 & 7.16 \\

&
&
Clean w/o Guide
& 48.88 & 45.75 & 44.31 & 3.13 & 4.57 \\

&
&
Combined w/ Guide
& 59.07 & 54.58 & 50.40 & 4.49 & 8.67 \\

&
\SetCell[r=3]{m} Llama 3.1 8B IT
& Clean w/ Guide
& 57.58 & 55.23 & 54.57 & 1.65 & 3.01 \\

&
&
Clean w/o Guide
& 55.27 & 46.92 & 49.86 & 8.35 & 5.41 \\

&
&
Combined w/ Guide
& 64.95 & 66.39 & 61.60 & -1.44 & 3.35 \\

&
\SetCell[r=3]{m} Llama 3.2 1B IT
& Clean w/ Guide
& 54.64 & 48.07 & 48.38 & 6.57 & 6.26 \\

&
&
Clean w/o Guide
& 47.03 & 43.89 & 40.42 & 3.14 & 6.61 \\

&
&
Combined w/ Guide
& 56.37 & 59.33 & 51.93 & -2.96 & 4.44 \\

\SetCell[r=8]{m} Encoder-Only
& \SetCell[r=2]{m} BanglaBERT Base
& Clean
& 67.35 & 56.79 & 51.35 & 8.02 & 15.42 \\

&
&
Combined
& 68.49 & 68.58 & 52.02 & -0.09 & 16.47 \\

&
\SetCell[r=2]{m} BanglaBERT Large
& Clean
& 68.91 & 58.34 & 50.25 & 10.57 & 18.66 \\

&
&
Combined
& 68.84 & 69.38 & 52.41 & -0.54 & 16.43 \\

&
\SetCell[r=2]{m} XLM-R Base
& Clean
& 67.35 & 55.46 & 58.54 & 11.89 & 8.81 \\

&
&
Combined
& 68.29 & 66.20 & 58.80 & 2.09 & 9.49 \\

&
\SetCell[r=2]{m} XLM-R Large
& Clean
& 70.61 & 56.92 & 61.43 & 13.69 & 9.18 \\

&
&
Combined
& 71.16 & 69.34 & 62.33 & 1.82 & 8.83 \\

\end{tblr}
}
\caption{Macro-F1 performance of encoder and decoder architectures across clean and noisy test sets. $\Delta_{\mathrm{ASR}}$ and $\Delta_{\mathrm{Ortho}}$ denote the absolute performance degradation relative to the clean test set ($\text{Clean} - \text{Noisy}$). Smaller $\Delta$ values indicate greater robustness, while negative values indicate improved performance on the noisy test set compared with the clean baseline.}

\label{tab:main_results_grid}
\end{table*}

\subsection{Evaluation of Language Models}
\paragraph{LLMs exhibit greater robustness to noise than encoder-based models.}

To quantify robustness, we measure the performance degradation due to noise, denoted by $\Delta \mathrm{F1}$, as the difference between a model's Macro F1 score on clean test set ($\mathrm{F1}_{\text{Clean}}$) and its average performance under noisy conditions ($\mathrm{F1}_{\text{Noisy}}$). To balance the influence of real-world ASR transcription errors and simulated orthographic noise, we define $\mathrm{F1}_{\text{Noisy}}=\frac{1}{2}(\mathrm{F1}_{\text{ASR}}+\mathrm{F1}_{\text{Ortho-Avg}})$, where $\mathrm{F1}_{\text{Ortho-Avg}}$ denotes the average Macro F1 across all word corruption rates. The overall robustness drop is then computed as $\Delta \mathrm{F1}=\mathrm{F1}_{\text{Clean}}-\mathrm{F1}_{\text{Noisy}}$.

\begin{figure*}[t]
  \centering
    \includegraphics[width=\textwidth]{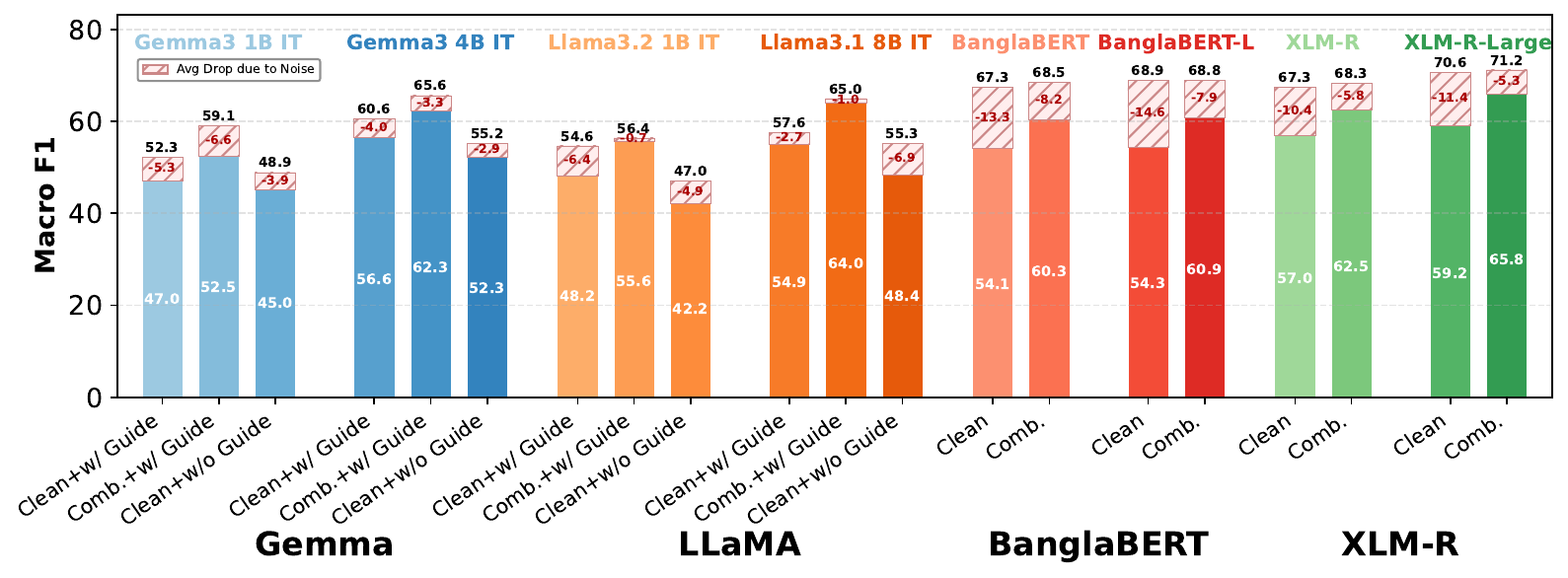}
    \caption{Decomposition of Macro-F1 performance across model architectures and training conditions into retained performance and average noise-induced degradation. The total height of each bar corresponds to the Clean Test Macro-F1 score. The solid region represents the average Macro-F1 retained across the ASR and simulated orthographic test sets, while the hatched region indicates the corresponding average performance drop relative to the clean test set.}
  \label{fig:overall_robustness}
  \end{figure*}

The results, as shown in Figure \ref{fig:overall_robustness}, reveal a clear distinction between encoder-based models and generative LLMs. Although encoder models achieve stronger performance on clean test set, they are substantially more sensitive to input corruption. For example, BanglaBERT large attains the highest clean-text Macro F1 score (68.9) but suffers a severe degradation of 14.6 points under noisy conditions. Similarly, XLM-R large experiences an average drop of 11.4 points. In contrast, generative LLMs exhibit considerably greater resilience. Llama 3.1 8B instruction tuned (w/ Guide) achieves a Macro F1 score of 57.6 on the clean test set while incurring an average performance drop of only 2.7 points under noisy conditions. Similarly, Gemma 3 4B IT (w/ Guide) experiences a degradation of merely 4.0 points. These results indicate a tradeoff between peak performance and robustness. Although encoder-based models trained exclusively on clean news text achieve higher Macro F1 scores on the clean test set, they experience substantial performance degradation under both orthographic and real-world ASR transcription noise. In contrast, generative LLMs exhibit considerably greater resilience and maintain more stable performance across noisy conditions.

\begin{figure*}[t]
  \centering
  \includegraphics[width=0.48\linewidth]{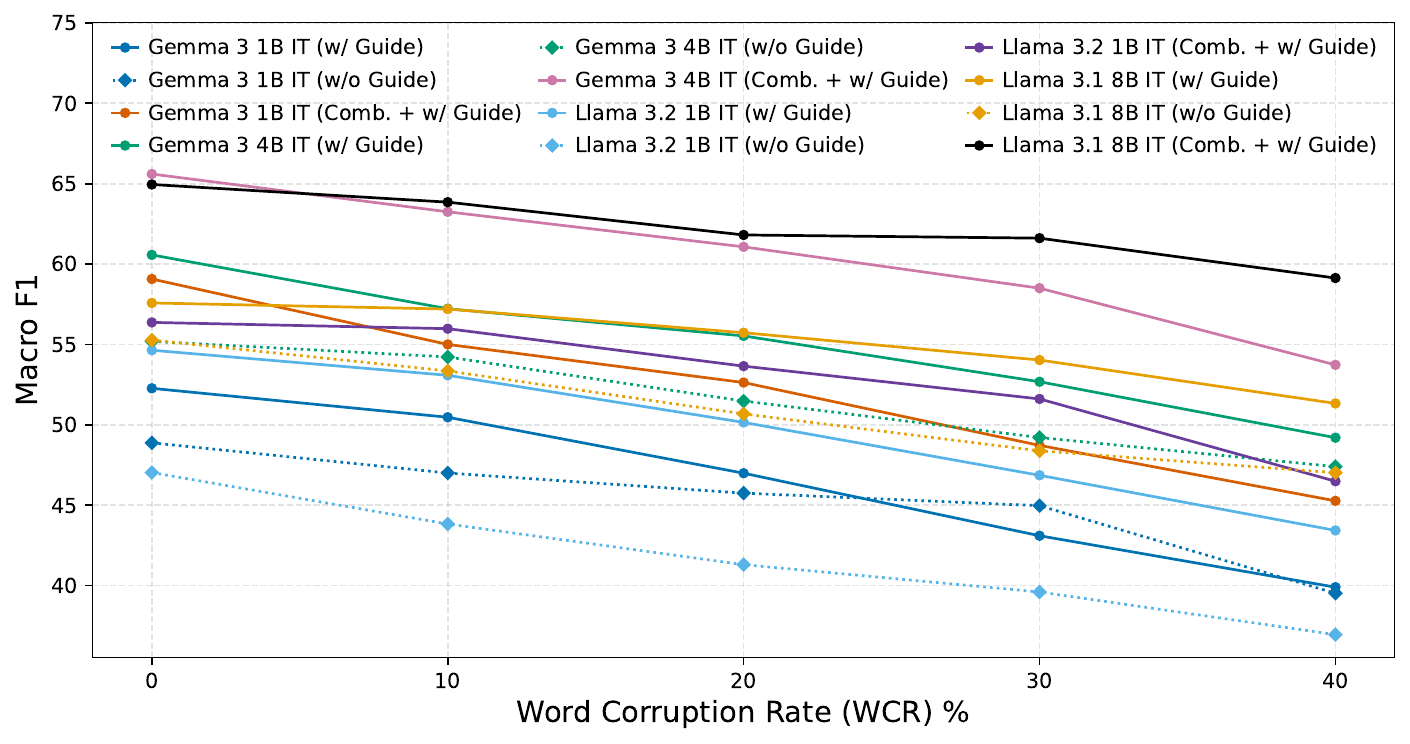} \hfill
  \includegraphics[width=0.48\linewidth]{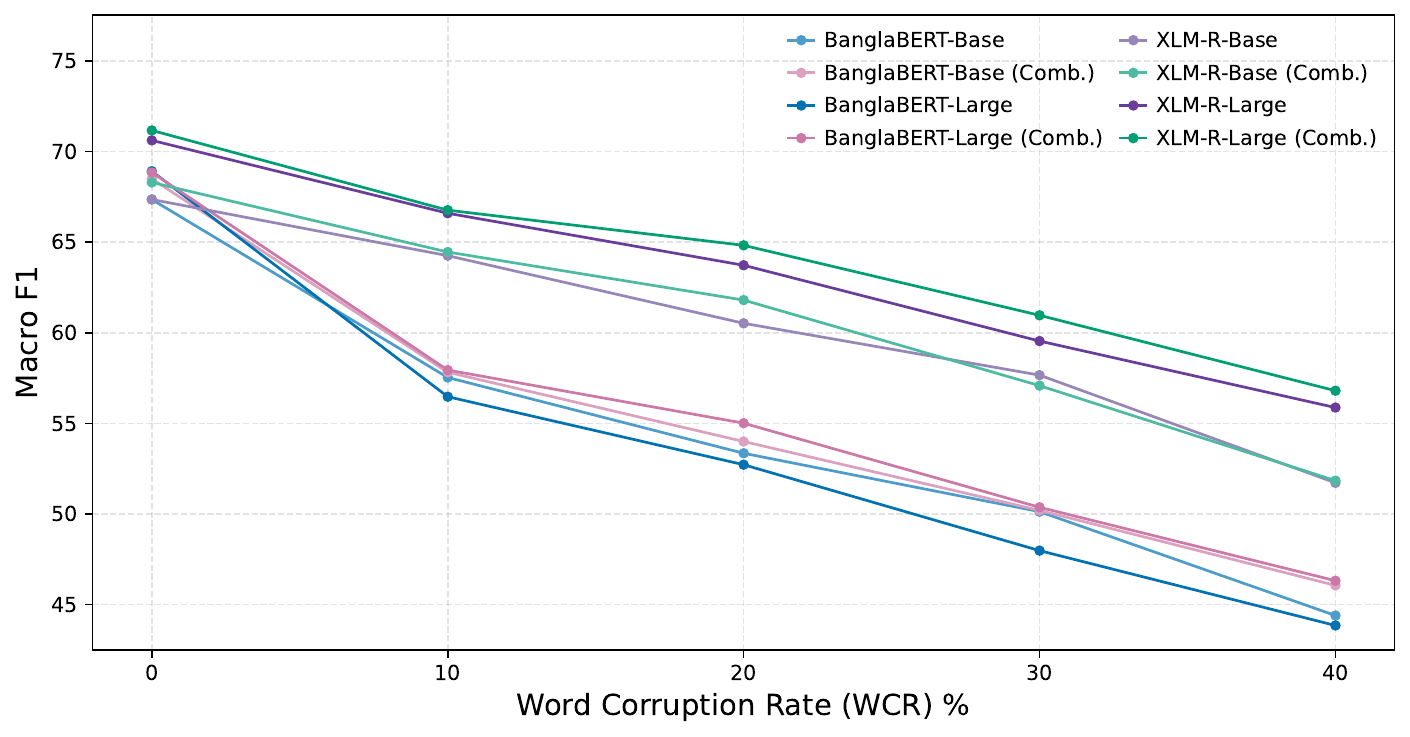}
  \caption{Comparison of performance degradation between Generative LLMs (left) and Encoder architectures (right) under varying levels of simulated orthographic noise.}
  \label{fig:wcr_degrade_comparison}
\end{figure*}

\paragraph{Instruction tuning with annotation guidelines improves performance but yields inconsistent reductions in clean-to-noisy performance drop}
As shown in Figure \ref{fig:wcr_degrade_comparison} and Table \ref{tab:main_results_grid}, across all evaluated LLM families, instruction tuning w/ Guide approach consistently improves performance on both clean and ASR test set and leads to higher Macro F1 score, even though the models are trained exclusively on clean news text. For example, w/ Guide approach increases the clean-test Macro F1 score of Gemma 3 4B from 55.2 to 60.6, with similar improvements observed across other models. However, the effect of annotation guideline on mitigating the relative performance drop ($\Delta \mathrm{F1}$) is inconsistent. For several models, w/ Guide approach slightly increase the magnitude of degradation; the average robustness drop for Gemma 3 4B widens from 2.9 points w/o Guide to 4.0 points with w/ Guide, while for Llama 3.2 1B, it increases from 4.9 to 6.4 points. In contrast, Llama 3.1 8B, benefits substantially from instruction tuning w/ Guide approach, with the average degradation decreasing from 6.9 points to only 2.6 points. These findings suggest that although annotation guidelines improve task understanding and overall performance, they only marginally reduce the performance degradation caused by noise.

\begin{figure}[t]
  \centering
    \includegraphics[width=\linewidth]{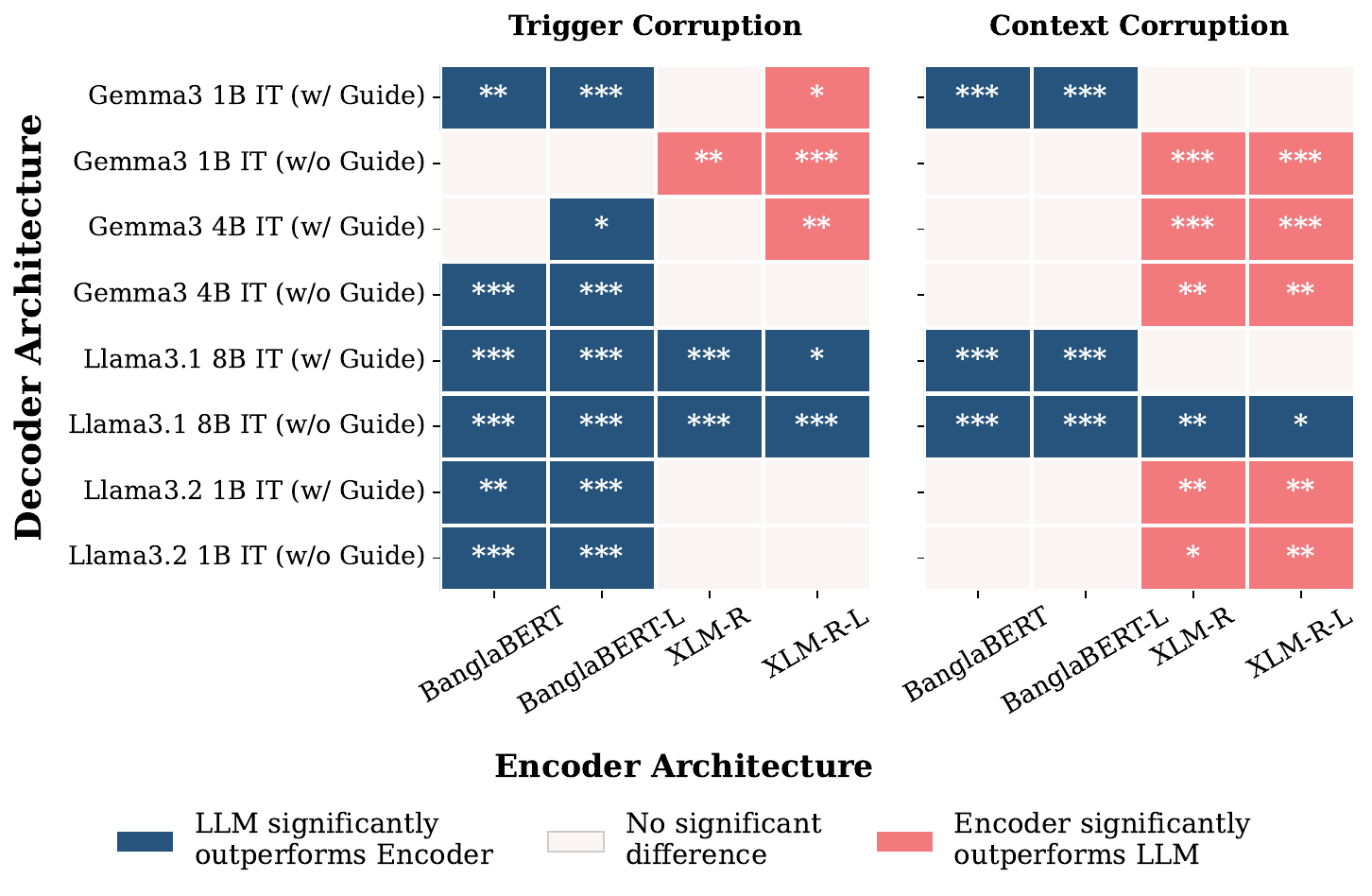}
    \caption{Pairwise McNemar's significance tests at $\mathrm{WCR}=40\%$ comparing decoder-only LLMs and encoder-only models on event detection. Results are reported separately for trigger corruption (left) and context corruption (right). Blue indicates that the LLM significantly outperforms the encoder, red indicates the converse, and white denotes no statistically significant difference (${}^{*}p<0.05$, ${}^{**}p<0.01$, ${}^{***}p<0.001$). All models are trained on clean news text and evaluated on previously unseen orthographically corrupted inputs.}
  \label{fig:trigger_context}
  \end{figure}

% \subsection{Analyzing Contextual Effects in Culturally Inappropriate Predictions}
\paragraph{Noise Combined training enhances encoder robustness}
Training on a mixture of clean and corrupted samples (Clean + Noisy ASR) improves robustness for both encoder-based models and generative LLMs. However, the magnitude of these gains differs considerably across architectures. For example, BanglaBERT Large's average performance drop ($\Delta \mathrm{F1}$) shrinks dramatically from 14.6 points to only 7.9 points. Similarly, for XLM-R shrinks degradation from 11.4 points to 5.3 points, with not notable increase in clean test performance for both. In contrast, generative LLMs benefit in both dimensions. For instance, Gemma 3 4B (w/ Guide) not only improves its clean-test performance but also reduces its average performance drop from 4.0 to 2.8 points, while Llama 3.1 8B (w/ Guide) simultaneously achieves higher clean-test score and further decreases its degradation from 2.6 to 1.7 points. These findings reveal a fundamental distinction between the two paradigms. Noise augmentation is particularly beneficial for encoder architectures, significantly narrowing the robustness gap between encoder models and generative LLMs.

\paragraph{Model Scaling Benefits Decoder-Only LLMs More Than Encoders}
An analysis of parameter scaling reveals a clear architectural divergence in robustness to noisy input conditions. Within decoder-only models, increasing model size consistently reduces the performance degradation induced by noise. For example, scaling from Llama 3.2 1B to Llama 3.1 8B decreases the average noise-induced Macro-F1 drop from 6.4 to 2.7 points under clean training w/ Guide. Similar trends are observed for the Gemma family, where the 4B model consistently incurs a smaller robustness penalty than its 1B counterpart. In contrast, scaling encoder models from Base to Large provides only marginal improvements in clean performance while failing to consistently improve robustness. Under clean training, BanglaBERT-Large and XLM-R-Large experience larger average performance drops than their Base counterparts (14.6 vs. 13.3 and 11.4 vs. 10.4 points, respectively), despite achieving higher clean-test performance. These results suggest that increased model capacity substantially improves the robustness of decoder-only architectures to noise, whereas simply scaling encoder architectures yields limited robustness gains. We hypothesize that larger decoder-only models better exploit semantic and contextual information to compensate for corrupted lexical forms, while larger encoder models remain comparatively sensitive to lexical perturbations despite their increased capacity.

\paragraph{Progressive orthographic noise induces monotonic performance decline, modulated by training strategy.}
Across increasing Word Corruption Rates (WCR), both generative LLMs and encoder-based architectures exhibit a monotonic decline in Macro F1 scores as simulated orthographic noise increases from 0\% to 40\% (Figure \ref{fig:wcr_degrade_comparison}). However, the severity of this degradation varies considerably across models and training strategies. Combined training consistently improves robustness by flattening the performance degradation curve, particularly at higher corruption levels. This effect is most pronounced for encoder-based models, where combined training substantially mitigates the steep decline observed in models trained exclusively on clean text. For generative LLMs, combined training provides an additional benefit by not only improving robustness but also increasing absolute performance across most corruption levels. Across all evaluated model families, w/ Guide variants consistently outperform their w/o Guide counterparts on both clean and corrupted inputs.

% Instruction guidance further enhances the performance of generative LLMs. Across all evaluated model families, guided variants consistently outperform their guideline-free counterparts on both clean and corrupted inputs. Although instruction guidance does not always yield substantial robustness gains, it establishes a higher performance baseline that remains largely preserved as noise increases. Overall, these findings indicate that while progressive orthographic corruption consistently degrades performance, both noise-augmented training and instruction guidance effectively mitigate its impact and lead to stronger performance under noisy conditions.

\paragraph{LLMs Excel Under Trigger Corruption, Encoders Under Context Corruption}
To complement the aggregate performance analysis, we perform paired statistical significance testing using McNemar's exact test at WCR = 40\%, the most challenging simulated orthographic corruption setting. Focusing on the highest corruption level allows us to determine whether performance differences remain statistically significant under severe noise. All models are trained on clean news text and evaluated on previously unseen orthographically corrupted inputs.

Predictions are aligned at the sentence level and partitioned according to whether corruption affects the event trigger itself or only the surrounding context. For decoder-only LLMs, generated trigger mentions are deterministically mapped back to their corresponding token spans in the original sentence through exact string matching, after which predictions are converted into \emph{(event type, start index, end index)} tuples identical to the encoder outputs.

For trigger corruption, a prediction is considered successful if the model correctly identifies the corrupted trigger span and assigns the correct event type despite the trigger word itself being orthographically corrupted. For context corruption, success requires correctly identifying the intact trigger span and its event type while remaining robust to orthographic noise introduced only in the surrounding context.

For each subset, predictions are converted into binary correct/incorrect outcomes based on strict trigger identification, where a prediction is considered correct only if both the trigger boundaries and event type exactly match the gold annotation. McNemar's exact test is then applied to the paired contingency table. Since McNemar's test evaluates only discordant prediction pairs, cases where both models are simultaneously correct or simultaneously incorrect do not contribute to the test statistic. Consequently, the reported significance reflects whether one model consistently succeeds on examples where the other fails. 
% We report the number of discordant pairs together with the corresponding significance levels to determine whether the observed differences between encoder-based models and decoder-only LLMs are statistically significant.
Overall, the results reveal a clear distinction between robustness to trigger corruption and context corruption. When the event trigger itself is orthographically corrupted, decoder-only LLMs, particularly the larger models, consistently outperform encoder-based architectures, indicating a significantly greater ability to correctly identify corrupted trigger mentions and assign the appropriate event type despite lexical perturbations. In contrast, when corruption is confined to the surrounding context while the trigger remains intact, the performance gap narrows considerably. Most model comparisons are no longer statistically significant, and among the significant cases, XLM-R frequently outperforms the smaller LLMs. This suggests that encoder-based models remain highly effective at exploiting contextual token representations when the trigger is preserved, whereas the principal advantage of decoder-only LLMs lies in maintaining accurate trigger identification under direct lexical corruption rather than in handling noisy contextual information alone.

\section{Conclusion}

In this paper, we presented the first comprehensive evaluation of language model robustness for Bangla event detection under both real-world ASR transcription errors and simulated orthographic noise. To facilitate this, we introduced a generalized Bangla news event ontology and released a benchmark comprising 9,979 human annotated sentences spanning clean news articles, ASR transcripts, and synthetic orthographic corruption. Through a systematic comparison of encoder-only and decoder-only architectures, we demonstrate a clear architectural trade-off: while encoder models achieve superior performance on clean text through precise trigger localization, they experience substantially greater degradation under lexical corruption. In contrast, decoder-only LLMs exhibit markedly stronger robustness, particularly when the event trigger itself is corrupted.

Our analyses further show that increasing model scale substantially improves the robustness of decoder-only LLMs, whereas scaling encoder architectures provides limited robustness gains despite improving clean-text performance. We also find that embedding annotation guidelines during instruction tuning establishes a higher performance baseline but yields inconsistent improvements in robustness to noisy inputs. In contrast, combined training on clean and noisy text consistently mitigates performance degradation, with particularly large gains for encoder-based models. Overall, our findings highlight the importance of evaluating event detection systems beyond clean-text benchmarks and suggest that decoder-only LLMs, together with combined training, provide a promising direction for robust event detection in real-world, low-resource settings.

\section*{Limitations}
Although we carefully designed our experiments to ensure a fair comparison between encoder-based models and decoder-only LLMs, our study has several limitations.

\paragraph{Strict Span-Based Evaluation.}
Our evaluation relies on the Strict Macro-F1 metric, which requires an exact match between the predicted and gold event trigger spans and event types. While this provides a rigorous assessment of trigger localization, it may disproportionately penalize generative LLMs. Encoder models are naturally formulated as sequence-labeling systems optimized for token-level BIO tagging, enabling precise prediction of trigger boundaries. In contrast, decoder-only LLMs generate event mentions autoregressively and occasionally produce semantically correct trigger mentions with slightly different surface boundaries (for instance, leaving out or adding a postposition or inflectional marker in Bangla). Such predictions are treated as incorrect under the strict evaluation protocol even when they capture the correct event. Consequently, the reported results should be interpreted as measuring exact trigger identification rather than semantic event understanding.
\paragraph{Domain and Task Scope.}
This study is restricted to the news domain, using datasets constructed from formal news articles and their same-domain ASR transcripts. Consequently, the reported robustness characteristics may not generalize to general-domain event detection, which may involve other event types and exhibit substantially different linguistic properties and noise distributions. Furthermore, our investigation is limited to event detection, namely identifying event trigger mentions and classifying their event types, as annotating event arguments, participant roles, temporal expressions, and other event attributes would require substantial annotation effort and computational resources. Future research should investigate whether the robustness trends observed for event detection extend to argument extraction and complete event extraction pipelines.

\section*{Ethical Considerations}

Annotators were compensated for their work at a rate consistent with local standards for similar annotation tasks. Prior to participation, annotators were informed about the purpose of the study and how their contributions would be used in the dataset. All news articles and Automatic Speech Recognition (ASR) transcripts were collected from publicly available channels. The dataset was compiled strictly for non-commercial, academic research purposes under standard fair use principles. Any named entities or individuals present in the text are public figures or subjects of public record, and no sensitive Personally Identifiable Information (PII) of private citizens was targeted or exposed.

\bibliography{custom}

\begin{thebibliography}{40}
\providecommand{\natexlab}[1]{#1}

\bibitem[{Ahn(2006)}]{ahnStagesEventExtraction2006}
David Ahn. 2006.
\newblock The stages of event extraction.
\newblock In \emph{Proceedings of the {{Workshop}} on {{Annotating}} and {{Reasoning}} about {{Time}} and {{Events}}}, pages 1--8, Sydney, Australia. Association for Computational Linguistics.

\bibitem[{Al~Monsur et~al.(2026)Al~Monsur, Bommisetty, and Kim}]{alMonsurEventDetectionContextAware2026}
Abdullah Al~Monsur, Nitesh~Vamshi Bommisetty, and Gene~Louis Kim. 2026.
\newblock \href {https://doi.org/10.18653/v1/2026.findings-eacl.314} {Event {{Detection}} with a {{Context-Aware Encoder}} and {{LoRA}} for {{Improved Performance}} on {{Long-Tailed Classes}}}.
\newblock In \emph{Findings of the {{Association}} for {{Computational Linguistics}}: {{EACL}} 2026}, pages 5985--6003, Rabat, Morocco. Association for Computational Linguistics.

\bibitem[{Ali~Khandokar et~al.(2025)Ali~Khandokar, All~Tanvir, Saddam Hossain~Mukta, and Shatabda}]{alikhandokarTemporalDemographicGeographical2025}
Iftakhar Ali~Khandokar, Abdullah All~Tanvir, {\relax Md}.~Saddam Hossain~Mukta, and Swakkhar Shatabda. 2025.
\newblock \href {https://doi.org/10.1007/s44230-025-00092-8} {Temporal, {{Demographic}}, and {{Geographical Analysis}} of {{Violent Events}} in {{Bangla News Media Using NLP Techniques}}}.
\newblock \emph{Human-Centric Intelligent Systems}, 5(1):90--102.

\bibitem[{Bhattacharjee et~al.(2022)Bhattacharjee, Hasan, Ahmad, Mubasshir, Islam, Iqbal, Rahman, and Shahriyar}]{bhattacharjeeBanglaBERTLanguageModel2022}
Abhik Bhattacharjee, Tahmid Hasan, Wasi Ahmad, Kazi~Samin Mubasshir, Md~Saiful Islam, Anindya Iqbal, M.~Sohel Rahman, and Rifat Shahriyar. 2022.
\newblock \href {https://doi.org/10.18653/v1/2022.findings-naacl.98} {{{BanglaBERT}}: {{Language Model Pretraining}} and {{Benchmarks}} for {{Low-Resource Language Understanding Evaluation}} in {{Bangla}}}.
\newblock In \emph{Findings of the {{Association}} for {{Computational Linguistics}}: {{NAACL}} 2022}, pages 1318--1327, Seattle, United States. Association for Computational Linguistics.

\bibitem[{Conneau et~al.(2020)Conneau, Khandelwal, Goyal, Chaudhary, Wenzek, Guzm{\'a}n, Grave, Ott, Zettlemoyer, and Stoyanov}]{conneauUnsupervisedCrosslingualRepresentation2020}
Alexis Conneau, Kartikay Khandelwal, Naman Goyal, Vishrav Chaudhary, Guillaume Wenzek, Francisco Guzm{\'a}n, Edouard Grave, Myle Ott, Luke Zettlemoyer, and Veselin Stoyanov. 2020.
\newblock \href {https://doi.org/10.18653/v1/2020.acl-main.747} {Unsupervised {{Cross-lingual Representation Learning}} at {{Scale}}}.
\newblock In \emph{Proceedings of the 58th {{Annual Meeting}} of the {{Association}} for {{Computational Linguistics}}}, pages 8440--8451, Online. Association for Computational Linguistics.

\bibitem[{Dave et~al.(2021)Dave, Gangopadhyay, Majumder, Bhattacharya, Sarkar, and Devi}]{daveFIRE2020EDNIL2021}
Bhargav Dave, Surupendu Gangopadhyay, Prasenjit Majumder, Pushpak Bhattacharya, Sudeshna Sarkar, and Sobha~Lalitha Devi. 2021.
\newblock \href {https://doi.org/10.1145/3441501.3441516} {{{FIRE}} 2020 {{EDNIL Track}}: {{Event Detection}} from {{News}} in {{Indian Languages}}}.
\newblock In \emph{Proceedings of the 12th {{Annual Meeting}} of the {{Forum}} for {{Information Retrieval Evaluation}}}, {{FIRE}} '20, pages 25--28, New York, NY, USA. Association for Computing Machinery.

\bibitem[{Dey et~al.(2021)Dey, Rahman, Mredula, Hosen, and Ra}]{deyUsingMachineLearning2021}
Noyon Dey, Md.~Sazzadur Rahman, Motahara~Sabah Mredula, A.~S. M.~Sanwar Hosen, and In-Ho Ra. 2021.
\newblock \href {https://doi.org/10.3390/electronics10192367} {Using {{Machine Learning}} to {{Detect Events}} on the {{Basis}} of {{Bengali}} and {{Banglish Facebook Posts}}}.
\newblock \emph{Electronics}, 10(19):2367.

\bibitem[{Dubey et~al.(2024)Dubey, Jauhri, and et~al.}]{dubey2024llama}
Abhimanyu Dubey, Abhinav Jauhri, and et~al. 2024.
\newblock The llama 3 herd of models.
\newblock \emph{arXiv preprint arXiv:2407.21783}.

\bibitem[{Ebner et~al.(2020)Ebner, Xia, Culkin, Rawlins, and Van~Durme}]{ebnerMultiSentenceArgumentLinking2020}
Seth Ebner, Patrick Xia, Ryan Culkin, Kyle Rawlins, and Benjamin Van~Durme. 2020.
\newblock \href {https://doi.org/10.18653/v1/2020.acl-main.718} {Multi-{{Sentence Argument Linking}}}.
\newblock In \emph{Proceedings of the 58th {{Annual Meeting}} of the {{Association}} for {{Computational Linguistics}}}, pages 8057--8077, Online. Association for Computational Linguistics.

\bibitem[{Elahi et~al.(2024)Elahi, Rahman, Shahriar, Sarker, Shawon, and Shahariar}]{elahiComparativeAnalysisNoise2024}
Kazi~Toufique Elahi, Tasnuva~Binte Rahman, Shakil Shahriar, Samir Sarker, Md~Tanvir~Rouf Shawon, and G.~M. Shahariar. 2024.
\newblock \href {https://doi.org/10.48550/arXiv.2401.14360} {A {{Comparative Analysis}} of {{Noise Reduction Methods}} in {{Sentiment Analysis}} on {{Noisy Bangla Texts}}}.
\newblock \emph{Preprint}, arXiv:2401.14360.

\bibitem[{{Gemma Team} et~al.(2025){Gemma Team}, Kamath, Ferret, Pathak, Vieillard, Merhej, Perrin, Matejovicova, Ram{\'e}, Rivi{\`e}re, Rouillard, Mesnard, Cideron, Grill, Ramos, Yvinec, Casbon, Pot, Penchev, Liu, Visin, Kenealy, Beyer, Zhai, Tsitsulin, {Busa-Fekete}, Feng, Sachdeva, Coleman, Gao, Mustafa, Barr, Parisotto, Tian, Eyal, Cherry, Peter, Sinopalnikov, Bhupatiraju, Agarwal, Kazemi, Malkin, Kumar, Vilar, Brusilovsky, Luo, Steiner, Friesen, Sharma, Sharma, Gilady, Goedeckemeyer, Saade, Kolesnikov, Bendebury, Abdagic, Vadi, Gy{\"o}rgy, Pinto, Das, Bapna, Miech, Yang, Paterson, Shenoy, Chakrabarti, Piot, Wu, Shahriari, Petrini, Chen, Lan, {Choquette-Choo}, Carey, Brick, Deutsch, Eisenbud, Cattle, Cheng, Paparas, Sreepathihalli, Reid, Tran, Zelle, Noland, Huizenga, Kharitonov, Liu, Amirkhanyan, Cameron, Hashemi, {Klimczak-Pluci{\'n}ska}, Singh, Mehta, Lehri, Hazimeh, Ballantyne, Szpektor, Nardini, {Pouget-Abadie}, Chan, Stanton, Wieting, Lai, Orbay, Fernandez, Newlan, Ji, Singh, Black, Yu, Hui,
  Vodrahalli, Greff, Qiu, Valentine, Coelho, Ritter, Hoffman, Watson, Chaturvedi, Moynihan, Ma, Babar, Noy, Byrd, Roy, Momchev, Chauhan, Bunyan, Botarda, Caron, Rubenstein, Culliton, Schmid, Sessa, Xu, Stanczyk, Tafti, Shivanna, Wu, Pan, Rokni, Willoughby, Vallu, Mullins, Jerome, Smoot, Girgin, Iqbal, Reddy, Sheth, P{\~o}der, Bhatnagar, Panyam, Eiger, Zhang, Liu, Yacovone, Liechty, Kalra, Evci, Misra, Roseberry, Feinberg, Kolesnikov, Han, Kwon, Chen, Chow, Zhu, Wei, Egyed, Cotruta, Giang, Kirk, Rao, Lo, Moreira, Martins, Sanseviero, Gonzalez, Gleicher, Warkentin, Mirrokni, Senter, Collins, Barral, Ghahramani, Hadsell, Matias, Sculley, Petrov, Fiedel, Shazeer, Vinyals, Dean, Hassabis, Kavukcuoglu, Farabet, Buchatskaya, Alayrac, Anil, Dmitry, {Lepikhin}, Borgeaud, Bachem, Joulin, Andreev, Hardin, Dadashi, and Hussenot}]{gemmateamGemma3Technical2025}
{Gemma Team}, Aishwarya Kamath, Johan Ferret, Shreya Pathak, Nino Vieillard, Ramona Merhej, Sarah Perrin, Tatiana Matejovicova, Alexandre Ram{\'e}, Morgane Rivi{\`e}re, Louis Rouillard, Thomas Mesnard, Geoffrey Cideron, Jean-bastien Grill, Sabela Ramos, Edouard Yvinec, Michelle Casbon, Etienne Pot, Ivo Penchev, and 193 others. 2025.
\newblock \href {https://doi.org/10.48550/ARXIV.2503.19786} {Gemma 3 {{Technical Report}}}.
\newblock \emph{arXiv preprint}.

\bibitem[{Hong et~al.(2011)Hong, Zhang, Ma, Yao, Zhou, and Zhu}]{hongUsingCrossEntityInference2011}
Yu~Hong, Jianfeng Zhang, Bin Ma, Jianmin Yao, Guodong Zhou, and Qiaoming Zhu. 2011.
\newblock Using {{Cross-Entity Inference}} to {{Improve Event Extraction}}.
\newblock In \emph{Proceedings of the 49th {{Annual Meeting}} of the {{Association}} for {{Computational Linguistics}}: {{Human Language Technologies}}}, pages 1127--1136, Portland, Oregon, USA. Association for Computational Linguistics.

\bibitem[{Hossain et~al.(2025)Hossain, {Sanjara}, Hossain, Chaki, Rahman, and Shawkat~Ali}]{hossainMaskNetEnhancingCrime2025}
Md.~Mithun Hossain, {Sanjara}, Md.~Shakil Hossain, Sudipto Chaki, Md.~Saifur Rahman, and A~B~M Shawkat~Ali. 2025.
\newblock \href {https://doi.org/10.1109/NCIM65934.2025.11160104} {{{MaskNet}}: {{Enhancing Crime Event Detection}} with {{Feature Masking}} and {{Dynamic Attention}}}.
\newblock In \emph{2025 2nd {{International Conference}} on {{Next-Generation Computing}}, {{IoT}} and {{Machine Learning}} ({{NCIM}})}, pages 1--6.

\bibitem[{Hu et~al.(2021)Hu, Shen, Wallis, {Allen-Zhu}, Li, Wang, Wang, and Chen}]{huLoRALowRankAdaptation2021a}
Edward~J. Hu, Yelong Shen, Phillip Wallis, Zeyuan {Allen-Zhu}, Yuanzhi Li, Shean Wang, Lu~Wang, and Weizhu Chen. 2021.
\newblock \href {https://doi.org/10.48550/arXiv.2106.09685} {{{LoRA}}: {{Low-Rank Adaptation}} of {{Large Language Models}}}.
\newblock \emph{Preprint}, arXiv:2106.09685.

\bibitem[{Huang et~al.(2024)Huang, Hsu, Parekh, Xie, Zhang, Natarajan, Chang, Peng, and Ji}]{huangTextEEBenchmarkReevaluation2024}
Kuan-Hao Huang, I-Hung Hsu, Tanmay Parekh, Zhiyu Xie, Zixuan Zhang, Prem Natarajan, Kai-Wei Chang, Nanyun Peng, and Heng Ji. 2024.
\newblock \href {https://doi.org/10.18653/v1/2024.findings-acl.760} {{{TextEE}}: {{Benchmark}}, {{Reevaluation}}, {{Reflections}}, and {{Future Challenges}} in {{Event Extraction}}}.
\newblock In \emph{Findings of the {{Association}} for {{Computational Linguistics}}: {{ACL}} 2024}, pages 12804--12825, Bangkok, Thailand. Association for Computational Linguistics.

\bibitem[{{HumanSignal}(2020)}]{label_studio}
{HumanSignal}. 2020.
\newblock {Label Studio}: Data labeling software.
\newblock Available at \url{https://labelstud.io}.

\bibitem[{Islam et~al.(2021)Islam, Kar, Islam, and Amin}]{islamSentNoBDatasetAnalysing2021}
Khondoker~Ittehadul Islam, Sudipta Kar, Md~Saiful Islam, and Mohammad~Ruhul Amin. 2021.
\newblock \href {https://doi.org/10.18653/v1/2021.findings-emnlp.278} {{{SentNoB}}: {{A Dataset}} for {{Analysing Sentiment}} on {{Noisy Bangla Texts}}}.
\newblock In \emph{Findings of the {{Association}} for {{Computational Linguistics}}: {{EMNLP}} 2021}, pages 3265--3271, Punta Cana, Dominican Republic. Association for Computational Linguistics.

\bibitem[{Khandokar et~al.(2020)Khandokar, Mamun, Chadni, Anas, and Shatabda}]{khandokarEventDetectionKnowledge2020a}
Iftakhar~Ali Khandokar, Imtiaz Mamun, Tasmia Ishrat~Alam Chadni, Zubair~Ahmed Anas, and Swakkhar Shatabda. 2020.
\newblock \href {https://doi.org/10.1109/ETCCE51779.2020.9350891} {Event {{Detection}} and {{Knowledge Mining}} from {{Unlabelled Bengali News Articles}}}.
\newblock In \emph{2020 {{Emerging Technology}} in {{Computing}}, {{Communication}} and {{Electronics}} ({{ETCCE}})}, pages 1--6, Bangladesh. IEEE.

\bibitem[{Kim et~al.(2009)Kim, Ohta, Pyysalo, Kano, and Tsujii}]{kimOverviewBioNLP09Shared2009}
Jin-Dong Kim, Tomoko Ohta, Sampo Pyysalo, Yoshinobu Kano, and Jun'ichi Tsujii. 2009.
\newblock Overview of {{BioNLP}}'09 {{Shared Task}} on {{Event Extraction}}.
\newblock In \emph{Proceedings of the {{BioNLP}} 2009 {{Workshop Companion Volume}} for {{Shared Task}}}, pages 1--9, Boulder, Colorado. Association for Computational Linguistics.

\bibitem[{Le and Nguyen(2021)}]{leFineGrainedEventTrigger2021}
Duong Le and Thien~Huu Nguyen. 2021.
\newblock \href {https://doi.org/10.18653/v1/2021.eacl-main.237} {Fine-{{Grained Event Trigger Detection}}}.
\newblock In \emph{Proceedings of the 16th {{Conference}} of the {{European Chapter}} of the {{Association}} for {{Computational Linguistics}}: {{Main Volume}}}, pages 2745--2752, Online. Association for Computational Linguistics.

\bibitem[{Leetaru and Schrodt(2013)}]{leetaru2013gdelt}
Kalev Leetaru and Philip Schrodt. 2013.
\newblock Gdelt: Global data on events, language, and tone, 1979-2012.
\newblock In \emph{International Studies Association Annual Conference}, San Francisco, CA.

\bibitem[{Li et~al.(2013)Li, Ji, and Huang}]{liJointEventExtraction2013}
Qi~Li, Heng Ji, and Liang Huang. 2013.
\newblock Joint {{Event Extraction}} via {{Structured Prediction}} with {{Global Features}}.
\newblock In \emph{Proceedings of the 51st {{Annual Meeting}} of the {{Association}} for {{Computational Linguistics}} ({{Volume}} 1: {{Long Papers}})}, pages 73--82, Sofia, Bulgaria. Association for Computational Linguistics.

\bibitem[{Loshchilov and Hutter(2019)}]{loshchilovDecoupledWeightDecay2019}
Ilya Loshchilov and Frank Hutter. 2019.
\newblock \href {https://doi.org/10.48550/arXiv.1711.05101} {Decoupled {{Weight Decay Regularization}}}.
\newblock \emph{Preprint}, arXiv:1711.05101.

\bibitem[{Nguyen et~al.(2021)Nguyen, Lai, Pouran Ben~Veyseh, and Nguyen}]{nguyenTrankitLightWeightTransformerbased2021}
Minh~Van Nguyen, Viet~Dac Lai, Amir Pouran Ben~Veyseh, and Thien~Huu Nguyen. 2021.
\newblock \href {https://doi.org/10.18653/v1/2021.eacl-demos.10} {Trankit: {{A Light-Weight Transformer-based Toolkit}} for {{Multilingual Natural Language Processing}}}.
\newblock In \emph{Proceedings of the 16th {{Conference}} of the {{European Chapter}} of the {{Association}} for {{Computational Linguistics}}: {{System Demonstrations}}}, pages 80--90, Online. Association for Computational Linguistics.

\bibitem[{Pang et~al.(2023)Pang, Cao, Ding, and Luo}]{pangGuidelineLearningInContext2023}
Chaoxu Pang, Yixuan Cao, Qiang Ding, and Ping Luo. 2023.
\newblock \href {https://doi.org/10.18653/v1/2023.emnlp-main.950} {Guideline {{Learning}} for {{In-Context Information Extraction}}}.
\newblock In \emph{Proceedings of the 2023 {{Conference}} on {{Empirical Methods}} in {{Natural Language Processing}}}, pages 15372--15389, Singapore. Association for Computational Linguistics.

\bibitem[{Patwardhan and Riloff(2009)}]{patwardhanUnifiedModelPhrasal2009}
Siddharth Patwardhan and Ellen Riloff. 2009.
\newblock A {{Unified Model}} of {{Phrasal}} and {{Sentential Evidence}} for {{Information Extraction}}.
\newblock In \emph{Proceedings of the 2009 {{Conference}} on {{Empirical Methods}} in {{Natural Language Processing}}}, pages 151--160, Singapore. Association for Computational Linguistics.

\bibitem[{Pouran Ben~Veyseh et~al.(2022)Pouran Ben~Veyseh, Nguyen, Dernoncourt, and Nguyen}]{pouranbenveysehMINIONLargeScaleDiverse2022}
Amir Pouran Ben~Veyseh, Minh~Van Nguyen, Franck Dernoncourt, and Thien Nguyen. 2022.
\newblock \href {https://doi.org/10.18653/v1/2022.naacl-main.166} {{{MINION}}: A {{Large-Scale}} and {{Diverse Dataset}} for {{Multilingual Event Detection}}}.
\newblock In \emph{Proceedings of the 2022 {{Conference}} of the {{North American Chapter}} of the {{Association}} for {{Computational Linguistics}}: {{Human Language Technologies}}}, pages 2286--2299, Seattle, United States. Association for Computational Linguistics.

\bibitem[{Saad et~al.(2024)Saad, Mahi, Salim, and Hossain}]{saadBanglaNewsArticle2024}
Asif~Mohammed Saad, Umme~Niraj Mahi, Md.~Shahidul Salim, and Sk~Imran Hossain. 2024.
\newblock \href {https://doi.org/10.1016/j.dib.2024.110874} {Bangla news article dataset}.
\newblock \emph{Data in Brief}, 57:110874.

\bibitem[{Sainz et~al.(2024)Sainz, {Garc{\'i}a-Ferrero}, Agerri, de~Lacalle, Rigau, and Agirre}]{sainzGoLLIEAnnotationGuidelines2024}
Oscar Sainz, Iker {Garc{\'i}a-Ferrero}, Rodrigo Agerri, Oier~Lopez de~Lacalle, German Rigau, and Eneko Agirre. 2024.
\newblock \href {https://doi.org/10.48550/arXiv.2310.03668} {{{GoLLIE}}: {{Annotation Guidelines}} improve {{Zero-Shot Information-Extraction}}}.
\newblock \emph{Preprint}, arXiv:2310.03668.

\bibitem[{Sechidis et~al.(2011)Sechidis, Tsoumakas, and Vlahavas}]{sechidis2011stratification}
Konstantinos Sechidis, Grigorios Tsoumakas, and Ioannis Vlahavas. 2011.
\newblock On the stratification of multi-label data.
\newblock \emph{Machine Learning and Knowledge Discovery in Databases}, pages 145--158.

\bibitem[{Sharif et~al.(2024)Sharif, Gatto, Basak, and Preum}]{sharifExplicitImplicitScattered2024}
Omar Sharif, Joseph Gatto, Madhusudan Basak, and Sarah~Masud Preum. 2024.
\newblock \href {https://doi.org/10.18653/v1/2024.emnlp-main.673} {Explicit, {{Implicit}}, and {{Scattered}}: {{Revisiting Event Extraction}} to {{Capture Complex Arguments}}}.
\newblock In \emph{Proceedings of the 2024 {{Conference}} on {{Empirical Methods}} in {{Natural Language Processing}}}, pages 12061--12081, Miami, Florida, USA. Association for Computational Linguistics.

\bibitem[{Sifat et~al.(2020)Sifat, Rahman, Rafsan, and Rahman}]{sifatSyntheticErrorDataset2020}
Md~Habibur~Rahman Sifat, Chowdhury~Rafeed Rahman, Mohammad Rafsan, and Md~Hasibur Rahman. 2020.
\newblock \href {https://doi.org/10.48550/arXiv.2003.03484} {Synthetic {{Error Dataset Generation Mimicking Bengali Writing Pattern}}}.
\newblock \emph{Preprint}, arXiv:2003.03484.

\bibitem[{Sims et~al.(2019)Sims, Park, and Bamman}]{simsLiteraryEventDetection2019}
Matthew Sims, Jong~Ho Park, and David Bamman. 2019.
\newblock \href {https://doi.org/10.18653/v1/P19-1353} {Literary {{Event Detection}}}.
\newblock In \emph{Proceedings of the 57th {{Annual Meeting}} of the {{Association}} for {{Computational Linguistics}}}, pages 3623--3634, Florence, Italy. Association for Computational Linguistics.

\bibitem[{Srivastava et~al.(2025)Srivastava, Pati, and Yao}]{srivastavaInstructionTuningLLMsEvent2025}
Saurabh Srivastava, Sweta Pati, and Ziyu Yao. 2025.
\newblock \href {https://doi.org/10.18653/v1/2025.findings-acl.677} {Instruction-{{Tuning LLMs}} for {{Event Extraction}} with {{Annotation Guidelines}}}.
\newblock In \emph{Findings of the {{Association}} for {{Computational Linguistics}}: {{ACL}} 2025}, pages 13055--13071, Vienna, Austria. Association for Computational Linguistics.

\bibitem[{Szymański and Kajdanowicz(2017)}]{pmlr-v74-szymański17a}
Piotr Szymański and Tomasz Kajdanowicz. 2017.
\newblock A network perspective on stratification of multi-label data.
\newblock In \emph{Proceedings of the First International Workshop on Learning with Imbalanced Domains: Theory and Applications}, volume~74 of \emph{Proceedings of Machine Learning Research}, pages 22--35, ECML-PKDD, Skopje, Macedonia. PMLR.

\bibitem[{Touileb et~al.(2024)Touileb, Murstad, M{\ae}hlum, Steskal, Storset, You, and {\O}vrelid}]{touilebEDENDatasetEvent2024}
Samia Touileb, Jeanett Murstad, Petter M{\ae}hlum, Lubos Steskal, Lilja~Charlotte Storset, Huiling You, and Lilja {\O}vrelid. 2024.
\newblock {{EDEN}}: {{A Dataset}} for {{Event Detection}} in {{Norwegian News}}.
\newblock In \emph{Proceedings of the 2024 {{Joint International Conference}} on {{Computational Linguistics}}, {{Language Resources}} and {{Evaluation}} ({{LREC-COLING}} 2024)}, pages 5495--5506, Torino, Italia. {ELRA and ICCL}.

\bibitem[{{Walker, Christopher} et~al.(2006){Walker, Christopher}, {Strassel, Stephanie}, {Medero, Julie}, and {Maeda, Kazuaki}}]{walkerchristopherACE2005Multilingual2006a}
{Walker, Christopher}, {Strassel, Stephanie}, {Medero, Julie}, and {Maeda, Kazuaki}. 2006.
\newblock \href {https://doi.org/10.35111/MWXC-VH88} {{{ACE}} 2005 {{Multilingual Training Corpus}}}.

\bibitem[{Wang et~al.(2020)Wang, Wang, Han, Jiang, Han, Liu, Li, Li, Lin, and Zhou}]{wangMAVENMassiveGeneral2020}
Xiaozhi Wang, Ziqi Wang, Xu~Han, Wangyi Jiang, Rong Han, Zhiyuan Liu, Juanzi Li, Peng Li, Yankai Lin, and Jie Zhou. 2020.
\newblock \href {https://doi.org/10.18653/v1/2020.emnlp-main.129} {{{MAVEN}}: {{A Massive General Domain Event Detection Dataset}}}.
\newblock In \emph{Proceedings of the 2020 {{Conference}} on {{Empirical Methods}} in {{Natural Language Processing}} ({{EMNLP}})}, pages 1652--1671, Online. Association for Computational Linguistics.

\bibitem[{Wang et~al.(2023)Wang, Li, and Ji}]{wangCode4StructCodeGeneration2023}
Xingyao Wang, Sha Li, and Heng Ji. 2023.
\newblock \href {https://doi.org/10.48550/arXiv.2210.12810} {{{Code4Struct}}: {{Code Generation}} for {{Few-Shot Event Structure Prediction}}}.
\newblock \emph{Preprint}, arXiv:2210.12810.

\bibitem[{Yao et~al.(2022)Yao, Xiao, Wang, Liu, Hou, Tu, Li, Liu, Shen, and Sun}]{yaoLEVENLargeScaleChinese2022}
Feng Yao, Chaojun Xiao, Xiaozhi Wang, Zhiyuan Liu, Lei Hou, Cunchao Tu, Juanzi Li, Yun Liu, Weixing Shen, and Maosong Sun. 2022.
\newblock \href {https://doi.org/10.18653/v1/2022.findings-acl.17} {{{LEVEN}}: {{A Large-Scale Chinese Legal Event Detection Dataset}}}.
\newblock In \emph{Findings of the {{Association}} for {{Computational Linguistics}}: {{ACL}} 2022}, pages 183--201, Dublin, Ireland. Association for Computational Linguistics.

\end{thebibliography}

\clearpage
\appendix

\section{Data Collection and Ontology Development}
\label{sec:ontology}
\subsection{Topic Coverage}

To ensure comparable topical coverage across clean text and ASR transcripts, we curated both corpora from the same set of news domains. For the clean text corpus, we selected articles from the BNAD \citep{saadBanglaNewsArticle2024} dataset, specifically using content from three major Bangladeshi newspapers: \textit{Dhaka Tribune Bangla}, \textit{The Daily Ittefaq}, and \textit{Daily Janakantha}s. We focused on seven major news domains: 
\begin{itemize}[leftmargin=*, itemsep=1pt]
    \item \bangla{রাজনীতি} (\textit{Politics})
    \item \bangla{অর্থনীতি} (\textit{Economy})
    \item \bangla{স্বাস্থ্য} (\textit{Health})
    \item \bangla{শিক্ষা} (\textit{Education})
    \item \bangla{প্রযুক্তি/টেক/বিজ্ঞান ও প্রযুক্তি} (\textit{Technology/Science})
    \item \bangla{বাণিজ্য} (\textit{Commerce})
    \item \bangla{চাকরি} (\textit{Jobs})
\end{itemize}

For the ASR corpus, we curated publicly available Bangla news broadcasts from the official YouTube channels of \textit{Jamuna TV}, \textit{Somoy TV}, and \textit{Independent Television}, and extracted their automatically generated Bangla speech transcripts. To mirror the clean text corpus, we collected broadcasts from category-specific playlists corresponding to the same seven news domains whenever available.

\subsection{Ontology Development}
To construct a generalized Bangla news event ontology, we analyzed a large-scale corpus of Bangladeshi news using the Global Database of Events, Language, and Tone (GDELT) through Google BigQuery and queried the Global Knowledge Graph (GKG) to extract historical event records related to Bangladesh from 2023 to 2025. Based on this corpus analysis, we identified event phenomena that are prevalent in Bangladeshi news and introduced locally relevant event types and subtypes to complement the widely adopted ACE 2005 ontology. This process ensured compatibility with established event types while capturing events that are underrepresented or absent in ACE 2005 but frequently occur in Bangladeshi news. The resulting ontology comprises 40 event subtypes organized into generalized event categories. The complete list of event types and subtypes, together with their source (ACE 2005 or GDELT-informed), is presented in Table~\ref{tab:event-types}.

\begin{figure}[t]
    \centering
    \includegraphics[width=0.5\linewidth]{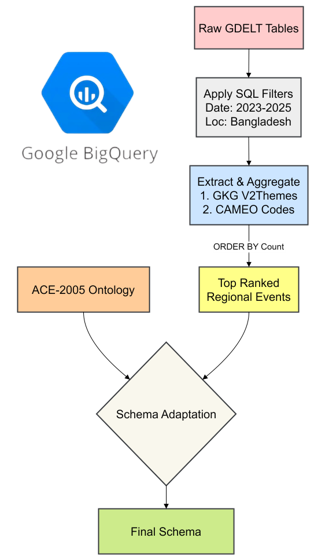}
    \caption{Overview of the ontology construction process. We analyze historical Bangladeshi news events from GDELT using Google BigQuery and the Global Knowledge Graph (GKG), adapt relevant event types, and integrate them with ACE 2005 to construct a generalized Bangla news event ontology.}
    \label{fig:GDELT}
\end{figure}

\begin{table*}[htbp]
\centering
\caption{Event types and sub types in the dataset}
\label{tab:event-types}
\begin{tabular}{|l|l|l|}
\hline
\textbf{Event Type} & \textbf{Sub Type} & \textbf{Source} \\
\hline
Contact & Meet & ACE 2005 \\
\hline
 & Phone-Write & ACE 2005 \\
\hline
Conflict & Attack & ACE 2005 \\
\hline
 & Demonstrate & ACE 2005 \\
\hline
Crime & Commit-Blue-Collar-Crime & GDELT \\
\hline
 & Commit-White-Collar-Crime &  GDELT\\
\hline
Disaster & Occur-Man-Made-Disaster & GDELT \\
\hline
 & Occur-Natural-Disaster & GDELT \\
\hline
Festival & Celebrate-Cultural-Festival & GDELT \\
\hline
 & Observe-Religious-Festival & GDELT \\
\hline
Governance & Approve & GDELT \\
\hline
 & Ban &  GDELT\\
\hline
 & Decide &  GDELT \\
\hline
 & Gazette & GDELT \\
\hline
 & Investigate &  GDELT\\
\hline
 & Reform & GDELT \\
\hline
 & Support & GDELT \\
\hline
Health & Outbreak & GDELT \\
\hline
 & Serve-Patients & GDELT \\
\hline
Justice & Arrest-Jail &  ACE 2005 \\
\hline
 & Charge-Indict & ACE 2005 \\
\hline
 & Deliver-Verdict & ACE 2005 \\
\hline
 & Sue & ACE 2005 \\
\hline
 & Trial-Hearing & ACE 2005 \\
\hline
Life & Die & ACE 2005 \\
\hline
 & Injure &  ACE 2005\\
\hline
Movement & Transport & ACE 2005 \\
\hline
Personnel & Elect & ACE 2005 \\
\hline
 & End-Position & ACE 2005 \\
\hline
 & Start-Position & ACE 2005 \\
\hline
Socio-Economy & Disrupt & GDELT  \\
\hline
 & Graduate & GDELT \\
\hline
 & Grow & GDELT \\
\hline
 & Import-Export & GDELT \\
\hline
 & Invest & GDELT \\
\hline
 & Recruit & GDELT \\
\hline
 & Trade & GDELT \\
\hline
Technology & Launch-Service & GDELT \\
\hline
Transaction & Transfer-Money & ACE 2005 \\
\hline
 & Transfer-Ownership & ACE 2005 \\
\hline
\end{tabular}
\end{table*}

\begin{figure*}[t]
    \centering
    \includegraphics[width=\textwidth]{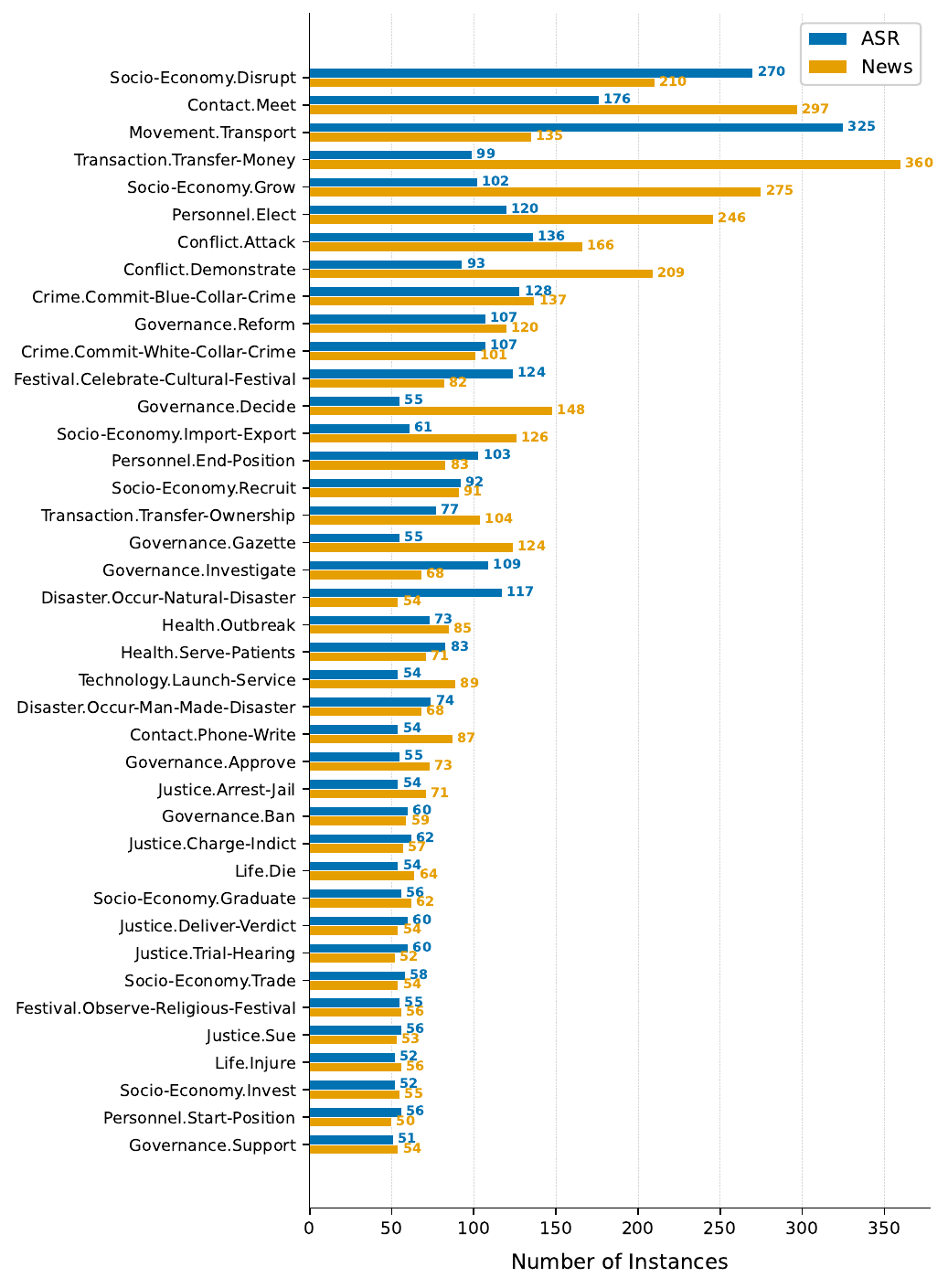}
    \caption{Distribution of event instances across the 40 event subtypes in the News and ASR corpora. Both corpora exhibit similar long-tail distributions, reflecting our topic-balancing effort despite the ASR corpus containing naturally occurring transcription artifacts and out-of-distribution vocabulary.}
    \label{fig:event_count}
\end{figure*}

\clearpage
% \subsection{Annotation Details}
\label{sec:annotate}
% \section{Evaluation Methodology}
% \label{sec:eval}
\section{Prompt Design and Model Training}
\subsection{Instruction Tuning Prompt Formats}
\label{sec:prompt_formats}

The following example illustrates the structured code generation prompt under the \textbf{w/ Guide} setting, where fine-grained annotation guidelines (event definitions, trigger examples, and boundary constraints) are embedded directly within the schema representation as Python docstrings. English translations are provided in brackets for reference.

\begin{tcolorbox}[colback=green!5!white,colframe=green!60!black,title=Instruction Tuning Prompt (w/ Guide)]
\begin{lstlisting}[breaklines=true]
# Instruction
Extract events according to the schema.

# Schema
@dataclass
class Injure(Life):
    """
    (*@\parbox{0.85\linewidth}{\bangla{এই ইভেন্টটি তখন ঘটে যখন অনিচ্ছাকৃত/স্বেচ্ছায়/দুর্ঘটনাবশত কোন ব্যক্তি শারীরিকভাবে আঘাত পান।}\\ \textcolor{transgray}{(This event occurs when a person is physically injured unintentionally/voluntarily/accidentally.)}}@*)
    
    (*@\parbox{0.85\linewidth}{\bangla{উদাহরণ ট্রিগার: আঘাত, ক্ষতি, আহত, অঙ্গহানি ইত্যাদি।}\\ \textcolor{transgray}{(Example triggers: injure, harm, injured, dismemberment, etc.)}}@*)

    (*@\bangla{সতর্কতা:} \textcolor{transgray}{(Note:)}@*)
    (*@\parbox{0.85\linewidth}{\bangla{- আঘাতপ্রাপ্ত ব্যক্তি মারা গেলে তা এই ইভেন্টের অন্তর্ভুক্ত হবে না।}\\ \textcolor{transgray}{(If the injured person dies, it will not be included in this event.)}}@*)
    """
    mention: str

# Sentence
text = "(*@\bangla{আহত শিক্ষার্থীকে উদ্ধার করা সম্ভব হয়েছে।}@*)"
       (*@\textcolor{transgray}{(The injured student was successfully rescued.)}@*)

result =

# Output
[
    Injure(mention="(*@\bangla{আহত}@*)") (*@\textcolor{transgray}{("injured")}@*)
]
\end{lstlisting}
\end{tcolorbox}

\begin{tcolorbox}[colback=gray!5!white,colframe=gray!60!black,title=Instruction Tuning Prompt (w/o Guide)]
\begin{lstlisting}[breaklines=true]
# Instruction
Extract events according to the schema.

# Schema
@dataclass
class Attack(Conflict):
    mention: str

# Sentence
text = "(*@\bangla{খুলনা বিশ্ববিদ্যালয়ে শিক্ষার্থী নির্যাতনের অভিযোগে শিক্ষার্থীর সীট বাতিল করেছে হল কর্তৃপক্ষ}@*)"
       (*@\textcolor{transgray}{(The hall authority has cancelled the seat of a student on allegations of student torture at Khulna University.")}@*)

result =

# Output
[
    Attack(mention="(*@\bangla{নির্যাতনের}@*)"),  (*@\textcolor{transgray}{("torture")}@*)
    ChargeIndict(mention="(*@\bangla{অভিযোগে}@*)")(*@\textcolor{transgray}{("allegations")}@*)
]
\end{lstlisting}
\end{tcolorbox}

\subsection{Training Details}
\label{sec:train}
\paragraph{Encoders}
Encoder-only models are fine-tuned using full-parameter supervised 
fine-tuning with the AdamW optimizer \citep{loshchilovDecoupledWeightDecay2019}, 
a learning rate of $2\times10^{-5}$, weight decay of $0.01$, 
warmup over $10\%$ of training steps, and a maximum sequence 
length of 128 tokens. Training is conducted for up to 15 epochs 
with early stopping based on validation Macro-F1 with patience 3.

\paragraph{Decoders}Decoder-only LLMs are adapted via Low-Rank Adaptation 
\citep{huLoRALowRankAdaptation2021a} with rank $r=16$, 
scaling factor $\alpha=32$, and dropout $p=0$, applied to all linear projection layers. We use a uniform rank across all model 
sizes following \citet{alMonsurEventDetectionContextAware2026}, 
who demonstrate that ranks above 8 provide no additional benefit 
for event detection tasks. Models are trained for up to 5 epochs 
with a learning rate of $1\times10^{-4}$, cosine learning rate 
schedule, warmup ratio of $0.05$, effective batch size of 64 
via gradient accumulation, and early stopping on validation loss 
with patience 3. All experiments are conducted 
on a single NVIDIA A100 GPU.
\end{document}